\documentclass[lettersize,journal]{IEEEtran}
\usepackage{algorithmicx}
\usepackage{algorithm}
\usepackage{array}
\usepackage[caption=false,font=normalsize,labelfont=sf,textfont=sf]{subfig}
\usepackage{textcomp}
\usepackage{stfloats}
\usepackage{url}
\usepackage{verbatim}
\usepackage{graphicx}
\usepackage{cite}
\usepackage{booktabs}
\usepackage{amssymb}
\usepackage{amsmath,amsfonts,bm}
\usepackage{multirow}
\usepackage{csquotes}
\usepackage{paralist}
\usepackage{threeparttable}
\usepackage{caption}

\usepackage[normalem]{ulem}

\newcommand\eg{\emph{e.g.}} 
\newcommand\ie{\emph{i.e.}} 
 
\newcommand\etc{\emph{etc.}}
\newcommand\wrt{w.r.t.}

\DeclareMathOperator*{\argmax}{arg\,max}

\usepackage{algpseudocode}
\usepackage[pagebackref,breaklinks,colorlinks]{hyperref}

\begin{document}

\title{Transferable Attack for Semantic Segmentation}


\author{Mengqi He,~
Jing Zhang,~
Zhaoyuan Yang,~
Mingyi He,~
Nick Barnes,~
Yuchao Dai
\\
\IEEEcompsocitemizethanks{\IEEEcompsocthanksitem Mengqi He, Jing Zhang and Nick Barnes are with School of Computing, Australian National University, Canberra, Australia. (mengqi.he@anu.edu.au, zjnwpu@gmail.com, nick.barnes@anu.edu.au)
\IEEEcompsocthanksitem Zhaoyuan Yang is with GE, America. (zhaoyuan.yang@ge.com)
\IEEEcompsocthanksitem Mingyi He and Yuchao Dai are with School of Electronics and Information, Northwestern Polytechnical University, Xi'an, China. (myhe@nwpu.edu.cn, daiyuchao@gmail.com)

}
}

\markboth{Journal of \LaTeX\ Class Files,~Vol.~14, No.~8, August~2021}%
{Shell \MakeLowercase{\textit{et al.}}: A Sample Article Using IEEEtran.cls for IEEE Journals}

\twocolumn[{%
\renewcommand\twocolumn[1][]{#1}%
\maketitle
\begin{center}
    \centering
  \begin{center}
  \begin{tabular}{{c@{ } c@{ } c@{ }}}
{\includegraphics[width=0.32\linewidth]{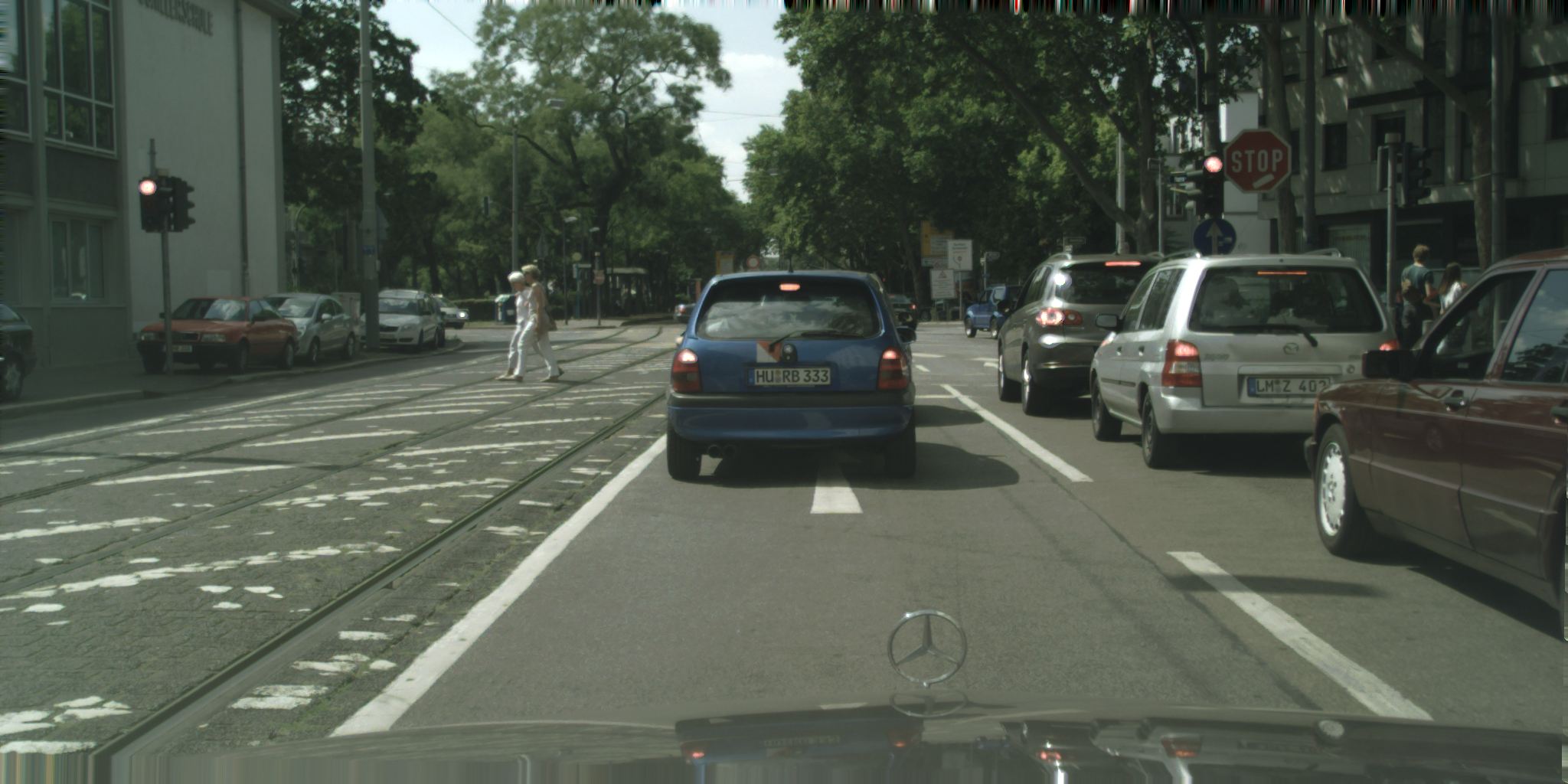}}&  
{\includegraphics[width=0.32\linewidth]{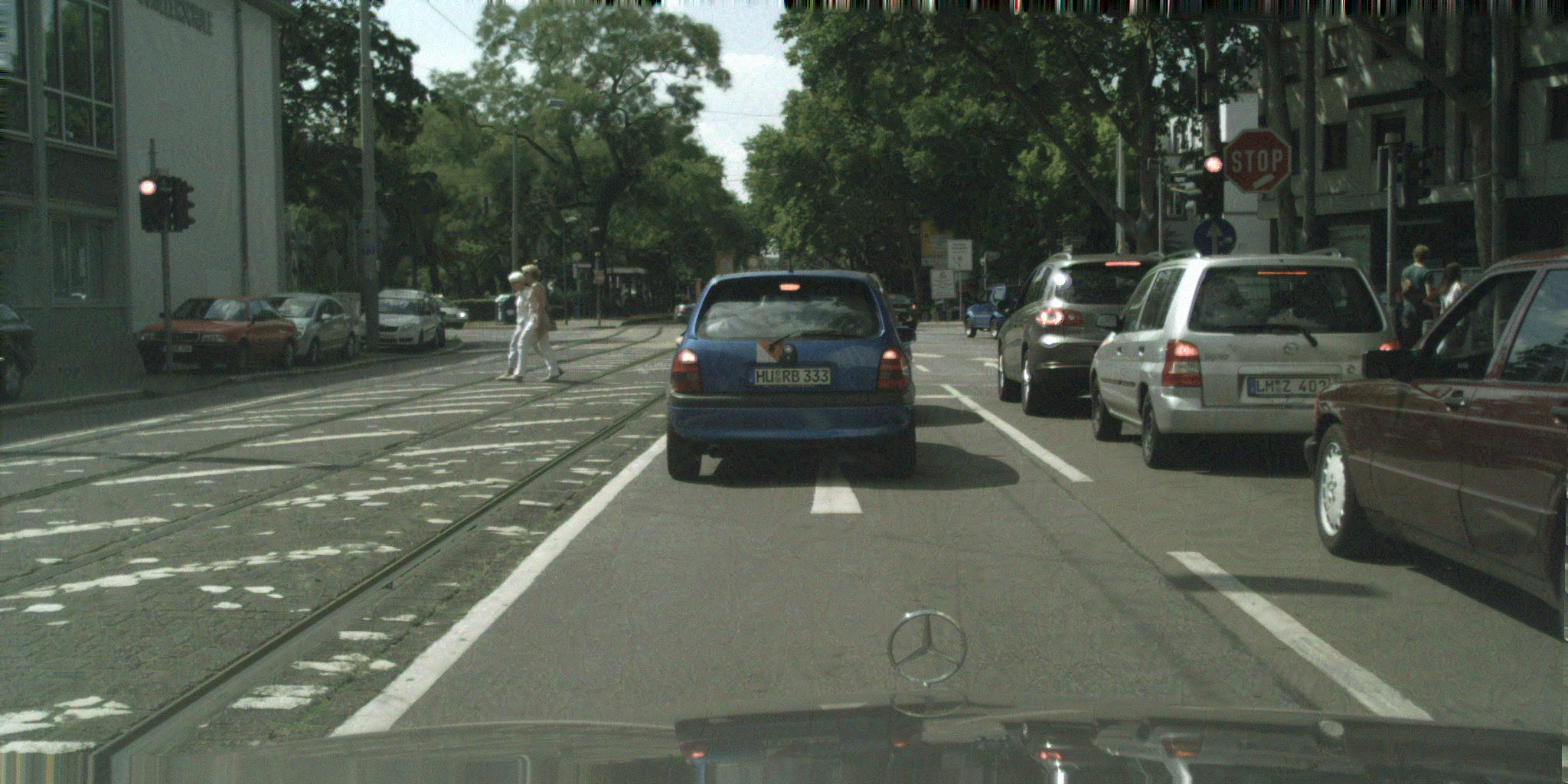}}&  
{\includegraphics[width=0.32\linewidth]{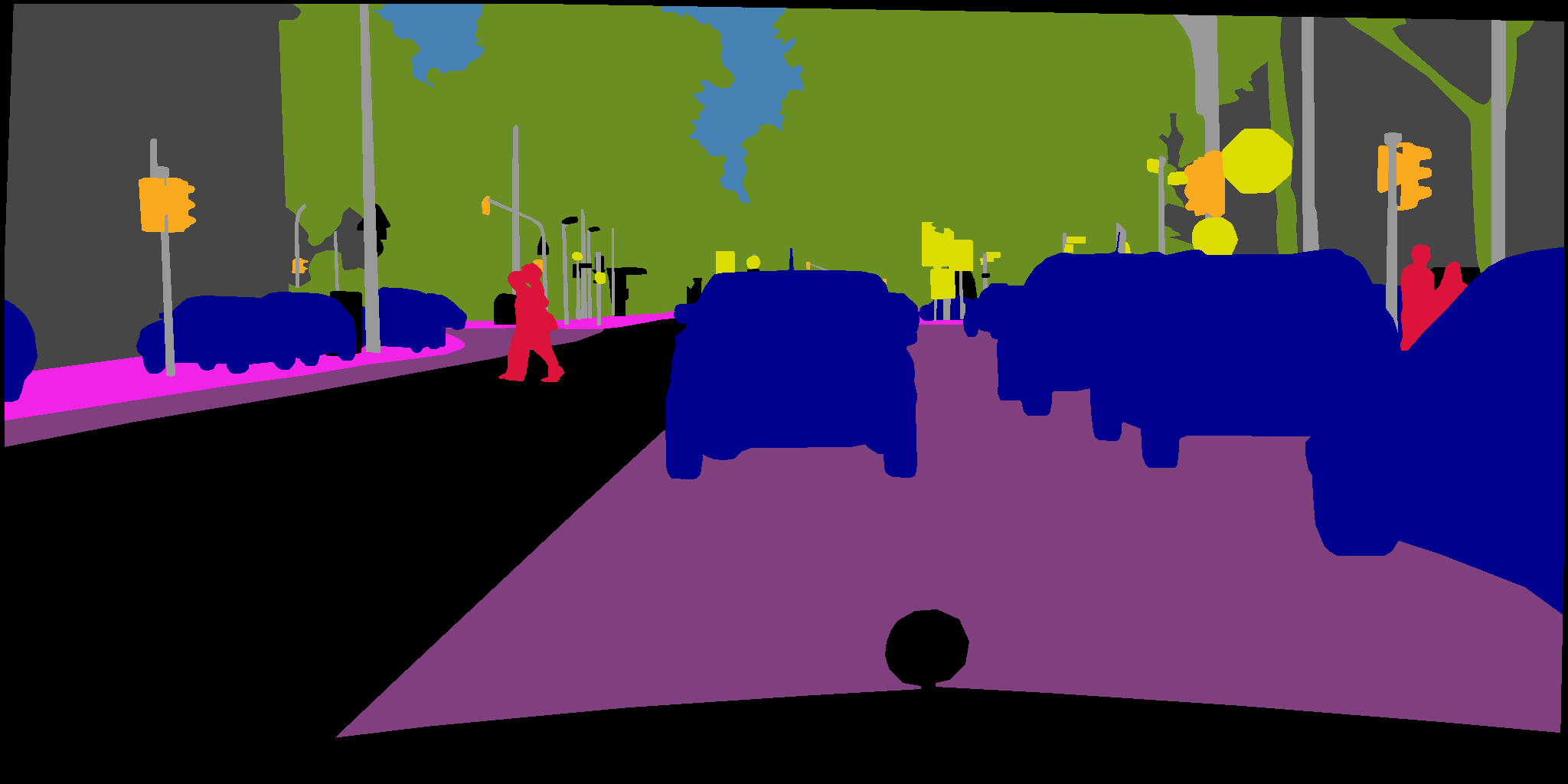}}\\ 
\footnotesize{Image}&\footnotesize{Adv}&\footnotesize{GT}\\
{\includegraphics[width=0.32\linewidth]{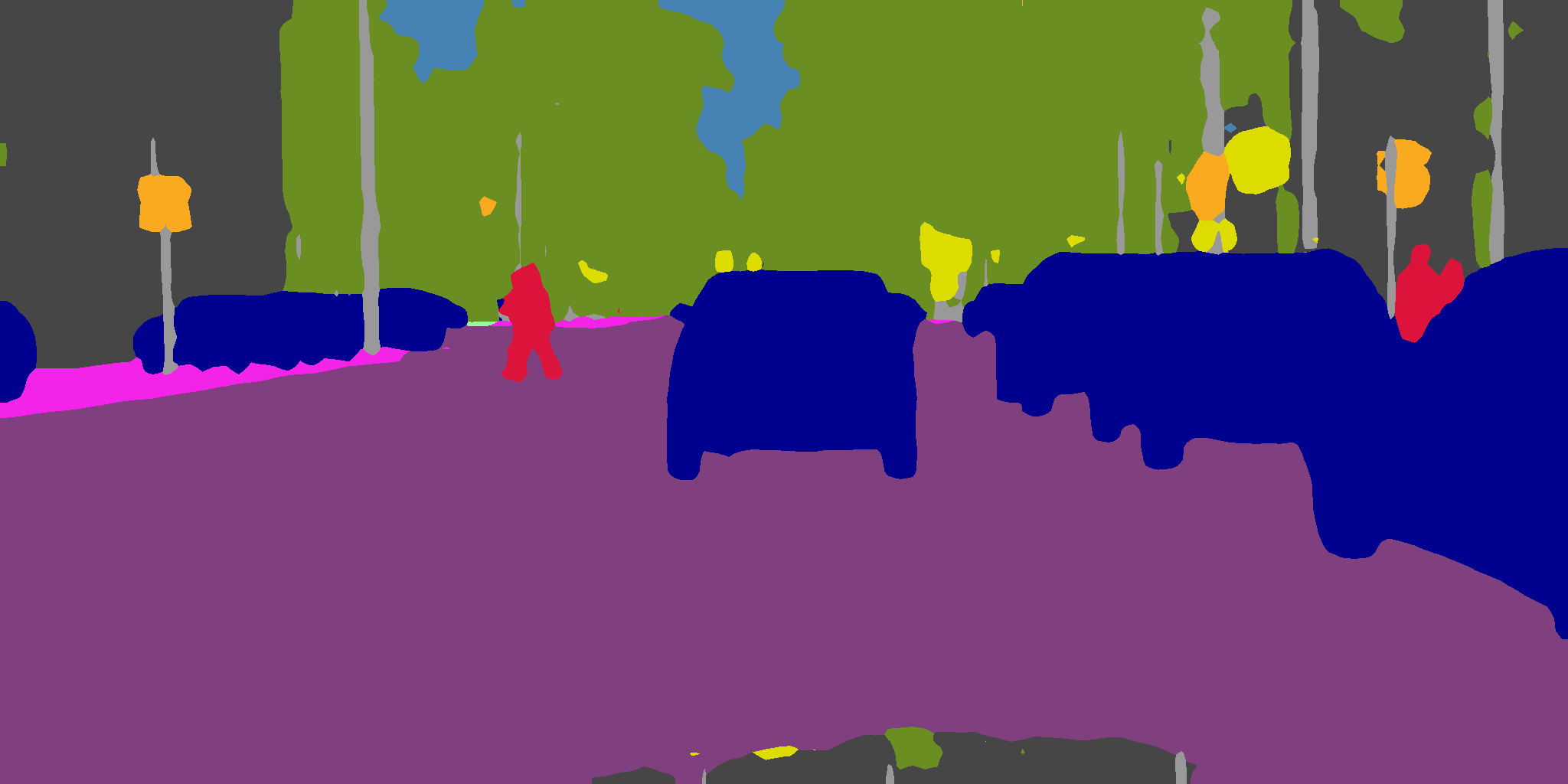}}&  
{\includegraphics[width=0.32\linewidth]{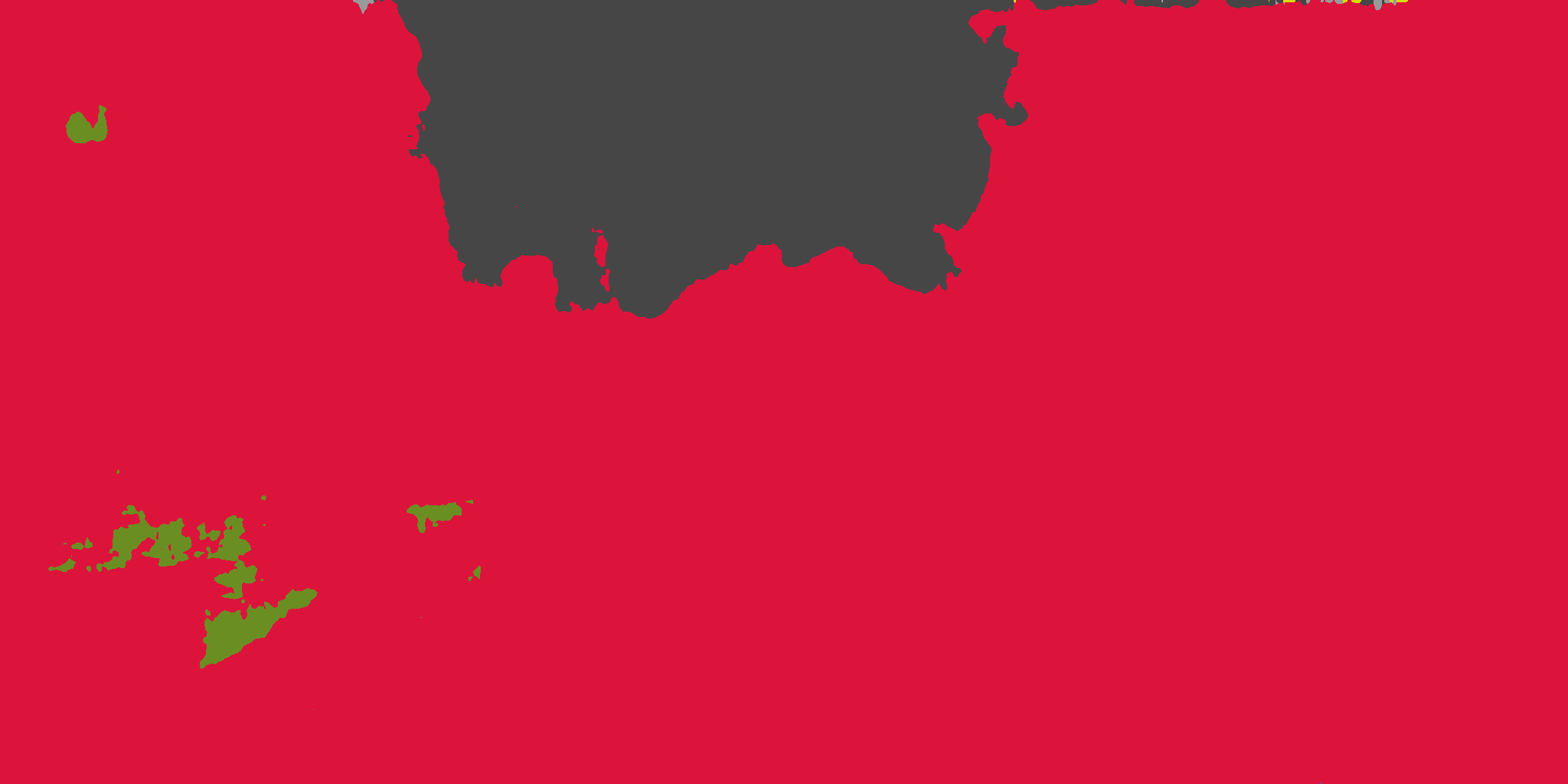}}&  
{\includegraphics[width=0.32\linewidth]{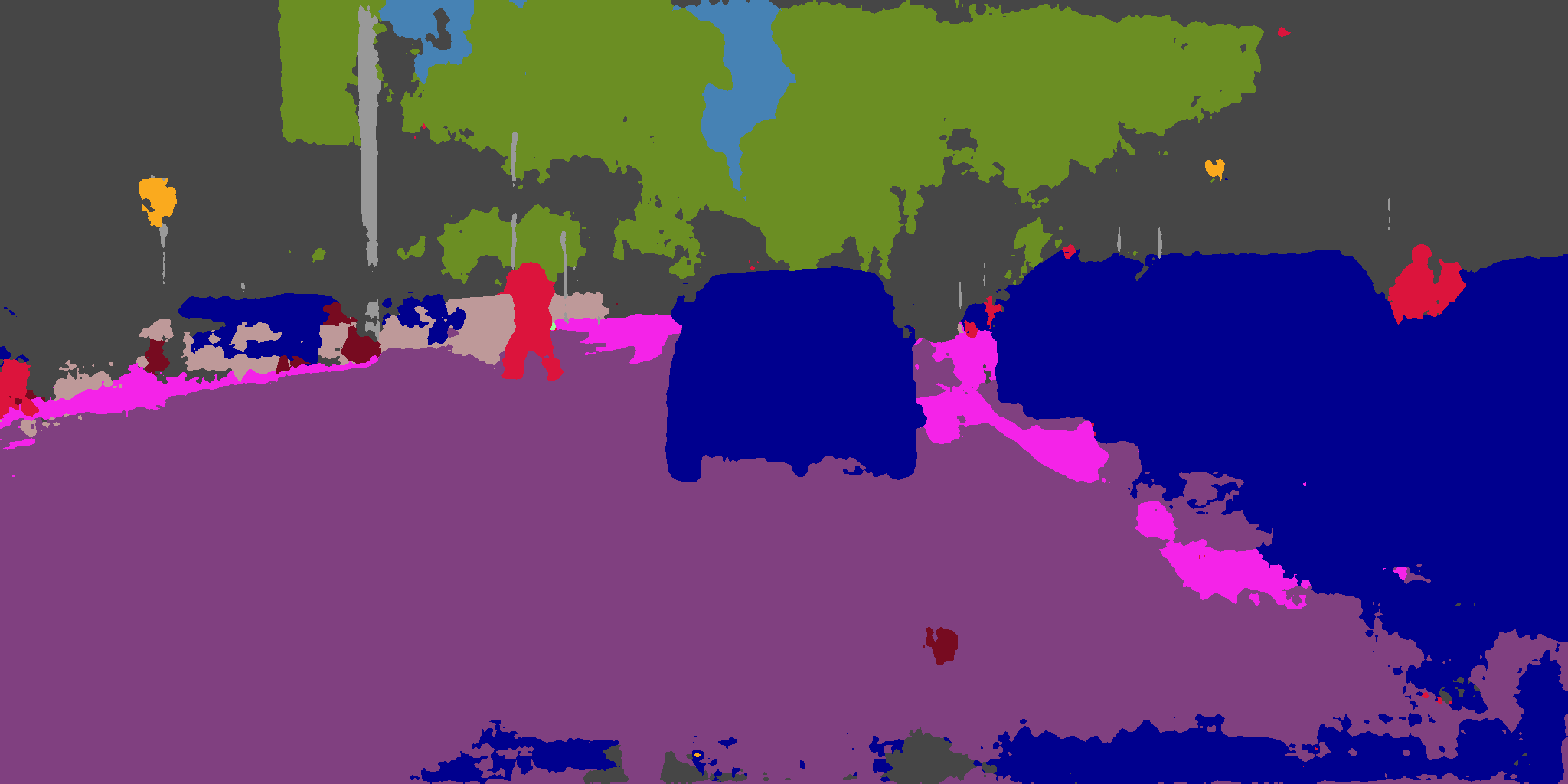}}\\
\footnotesize{Clean}&\footnotesize{Noise}&\footnotesize{Noise+}\\
   \end{tabular}
  \end{center}
    \captionof{figure}{
   Robustness of semantic segmentation models to adversarial attacks. Image: a clean RGB image from Cityscape dataset~\cite{Cordts2016Cityscapes}, Adv: the adversarial example generated from the source model Deeplabv3~\cite{chen2017rethinking} with PGD~\cite{PGD_attack} attack, GT: the ground truth,  Clean: Deeplabv3 output of the clean image; Noise: Deeplabv3 output of the adversarial example, Noise+: Deeplabv3+~\cite{chen2018encoder} output of the adversarial example from Deeplabv3~\cite{chen2017rethinking}. The better performance of \enquote{Noise+} compared with \enquote{Noise} indicates that the adversarial example from source model (\enquote{Deeplabv3}) fails to attack the target model (\enquote{Deeplabv3+}), explaining the necessity of studying transferable attack for semantic segmentation.
    } 
    \label{semantic_seg_towards_attack}
\end{center}%

}]

\begin{abstract}
We analysis performance of semantic segmentation models \wrt~adversarial attacks, and observe that the adversarial examples generated from a source model fail to attack the target models. \ie~The conventional attack methods, such as PGD~\cite{PGD_attack} and FGSM~\cite{Explaining_and_Harnessing_Adversarial_Examples}, do not transfer well to target models, making it necessary to study the transferable attacks, especially transferable attacks for semantic segmentation. We find two main factors to achieve transferable attack. Firstly,
the attack should come with effective data augmentation and translation-invariant features to deal with unseen models. Secondly, stabilized optimization strategies are needed to find the optimal attack direction. Based on the above observations, we propose an ensemble attack for semantic segmentation
to achieve more effective attacks with higher transferability. The source code and experimental results are publicly available via our project page: \url{https://github.com/anucvers/TASS}


\end{abstract}


\section{Introduction}
\IEEEPARstart{S}{emantic} segmentation~\cite{chen2022vision,yuan2019segmentation,yan2022lawin,mohan2021efficientps,cheng2020panoptic,zhang2021dcnas,li2019global,chen2017rethinking,chen2018encoder,long2015fully,lin2017refinenet,chen2016attention,acuna2019devil,girshick2014rich,chen2017deeplab,deng2019restricted,goodfellow2020generative,kingma2013auto,li2021semantic,souly2017semi,luc2016semantic,xue2018segan,zhaoa2021semantic} is widely studied.
Although performance is highly advanced,
we find the existing semantic segmentation models
are vulnerable to input noise~\cite{xie2017adversarial,arnab2018robustness}, where minor and invisible perturbations can ruin the prediction.
In Fig.~\ref{semantic_seg_towards_attack}, we show examples to explain model fragility with respect to adversarial attack, which clearly shows the invisible difference between
clean image (\enquote{Image}) and the adversarial example (\enquote{Adv}) can cause significant performance decrease (see \enquote{Clean} and \enquote{Noise} for predictions of \enquote{Image} and \enquote{Adv} from the source model, namely Deeplabv3~\cite{chen2017rethinking}, respectively), 
indicating vulnerability of semantic segmentation models to adversarial attacks.

Although existing attacks for semantic segmentation~\cite{gu2022segpgd,xie2017adversarial} can roughly be used to evaluate model robustness of semantic segmentation models,
we find that one main limitation of those techniques is that
the success rate of attacks for segmentation is poor when it transfers to different network structures, indicating poor transferability of the attacks.
In Fig.~\ref{semantic_seg_towards_attack}, we further
test effectiveness of adversarial example (\enquote{Adv}) on a target model with a different model structure, e.g. Deeplabv3+~\cite{chen2018encoder} in this case.
We find that the adversarial example from the
source model (\enquote{Deeplabv3}) fails to attack the target model (\enquote{Deeplabv3+}), which we claim as one main motivation for investigating of transferable attack.

Further, in black-box scenario, where target model weights are unknown, with transferable attack,
we could use a source model with known weights to generate the attack and transfer the attack to the target model with unknown weights. Also, for scenarios where the
target model is too large or computationally expensive to generate attack, if we could generate attack from a smaller model and transfer to larger model, computation cost can be significantly reduced.

With the above three main motivations, we work on transferable attacks for semantic segmentation. Firstly,
we extensively explore conventional image classification-based adversarial attacks and transferable attacks, and extend those solutions to semantic segmentation, a dense pixel-wise classification task. Secondly,
we observe three main attributes, which include proper data augmentation, gradient stabilization and network structure refinement to achieve effective transferable attacks, and
present ensemble attack
to achieve better transferable attacks. 
The ensemble attacks work because it realizes the data augmentation and the gradient stabilization at the same time.
Lastly, we consider the robustness of vision foundation models~\cite{kirillov2023segany} \wrt~transferable adversarial attack, further explaining the necessity of our investigation on the transferable attack for semantic segmentation.



\section{Related Work}

\noindent\textbf{Adversarial attack:} tries to generate invisible perturbations
to evaluate model robustness.
The generation of adversarial examples is an optimization problem. Define classification loss of a neural network as $\mathcal{L}$, parameters of a neural network as $\theta$, input data as $\mathbf{x}$ and its corresponding one-hot encoded label as $\mathbf{y}$. The objective of adversarial attacks~\cite{szegedy2013intriguing, Explaining_and_Harnessing_Adversarial_Examples, PGD_attack, Towards_Evaluating_Robustness_Neural_Networks, Ensemble_Adversarial_Training_Attacks_Defenses, DeepFool, szegedy2013intriguing} (Eq.~\ref{eq:atk_obj}) is to maximize the classification loss such that a classifier will make incorrect predictions. We define $||\cdot||_p$ as the $L_p$ norm. The norm of an adversarial perturbation $\delta_{adv}$ needs to be bounded by a small value $\delta_{t}$, otherwise the adversarial perturbation may change the semantic interpretation of an image. Once an adversarial perturbation is obtained, the adversarial example can be crafted as $\mathbf{x}_{adv} = \mathbf{x} + \delta_{adv}$, where
\begin{equation}
\label{eq:atk_obj}
\delta_{adv} = \argmax_{||\delta||_{p} \leq \delta_{t}} \mathcal{L}(\theta,\mathbf{x}+\delta,\mathbf{y}).
\end{equation}
Two typical adversarial attacks include
FGSM~\cite{Explaining_and_Harnessing_Adversarial_Examples} and PGD~\cite{PGD_attack}. The former
generates the perturbations based
on the opposite direction of the loss gradient or the direction for gradient ascent:
\begin{equation}
\label{eq_fgsm_attack}
    \delta_{adv} = \epsilon \text{sign} \left( \nabla_{\mathbf{x}} \mathcal{L}(\theta, \mathbf{x}, \mathbf{y}) \right),
\end{equation}
where $\epsilon $ is the perturbation rate, $\text{sign}$ is the sign function,
$\mathcal{L}(\theta, \mathbf{x}, \mathbf{y})$ is the loss function, and $\theta$ represents model parameters. FGSM~\cite{Explaining_and_Harnessing_Adversarial_Examples} works well for most simple scenarios. However, it might fail to generate accurate directions of adversarial examples due to its single-step scheme.
PGD~\cite{PGD_attack} is then introduced as a multi-step variant:
\begin{equation}
\label{eq_pgd_attack}
    \mathbf{x}_{adv}^{t+1} = \text{CLIP}(\mathbf{x}_{adv}^t + \alpha\text{sign} \left( \nabla_{\mathbf{x}_{adv}^t} \mathcal{L}(\theta, \mathbf{x}_{adv}^t, \mathbf{y}) \right)),
\end{equation}
where
$\alpha$ is the step size, 
$\text{CLIP}$ is the clip function that clip the output into the range $\mathbf{x}_{adv}^0=\mathbf{x}_{clean}+\mathcal{U}(-\epsilon,\epsilon)$ specified by the perturbation rate $\epsilon$.

\begin{figure}[tp]
   \begin{center}
   \begin{tabular}{c@{ } c@{ } c@{ }c@{ }}
{\includegraphics[width=0.230\linewidth]{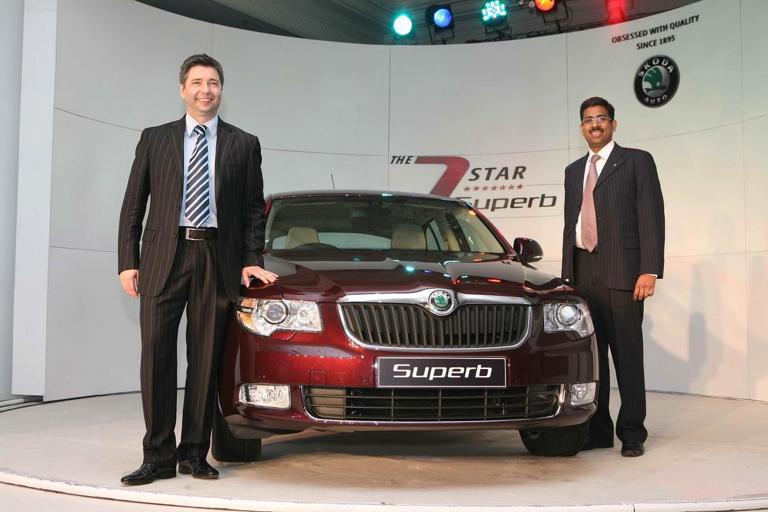}}&  
{\includegraphics[width=0.230\linewidth]{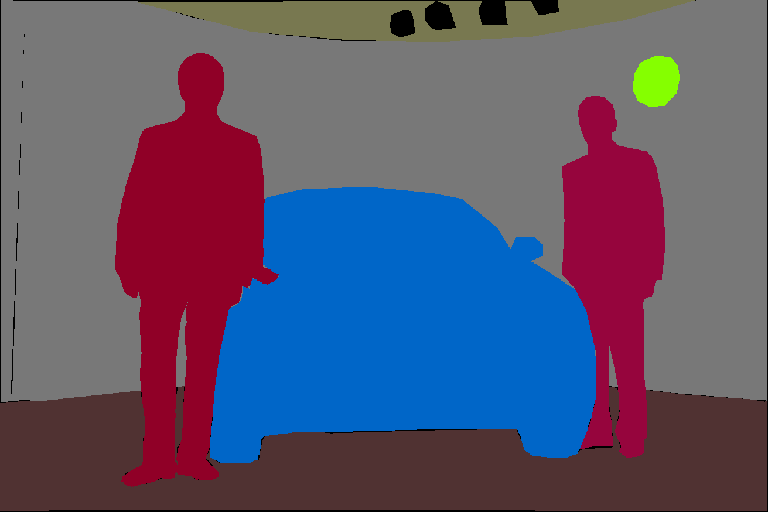}}&  
{\includegraphics[width=0.230\linewidth]{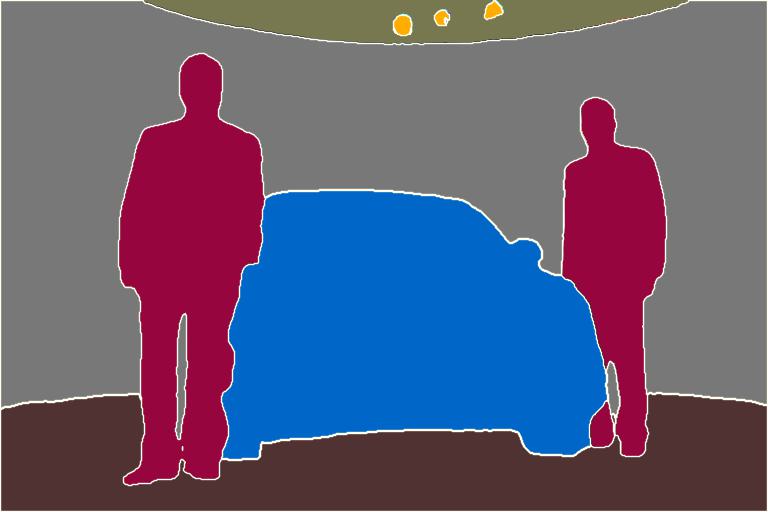}}&  
{\includegraphics[width=0.230\linewidth]{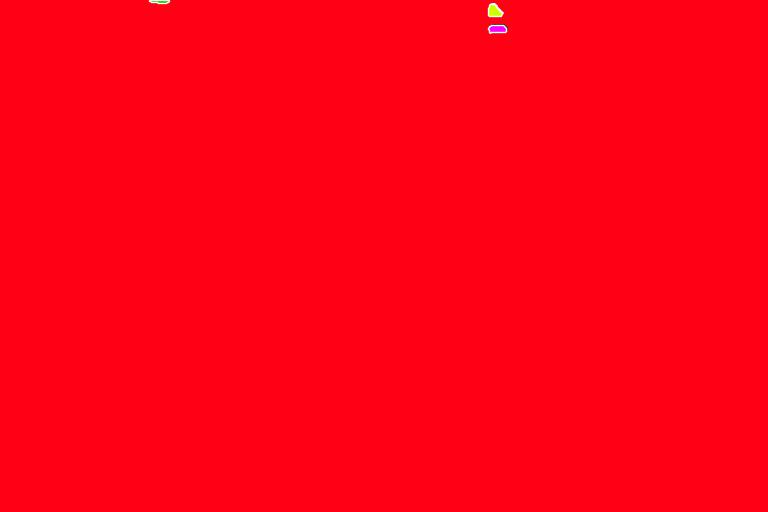}}\\ 
\footnotesize{Image}&\footnotesize{GT}&\footnotesize{Maskformer~\cite{cheng2021perpixel}}&\footnotesize{Attacked}\\
   \end{tabular}
   \end{center}
    \caption{Performance of transformer based semantic segmentation models \wrt~adversarial attack, where \enquote{Attacked} is the segmentation result of Maskformer~\cite{cheng2021perpixel} for the attacked image using PGD~\cite{PGD_attack} attack, where the model predicts every pixel to be the category of \enquote{towel}.
     } 
    \label{fig:segmentation_robustness_wrt_transformer}
\end{figure}

\noindent\textbf{Semantic segmentation models:}
With the pioneer work from~\cite{long2015fully}, semantic segmentation is widely studied, including both CNN based techniques
\cite{acuna2019devil,Takikawa_2019_ICCV,yuan2020segfix,Liang_2020_CVPR,chen2017deeplab,wang2018understanding,zhao2017pyramid,deng2019restricted,dai2017deformable,liu2015semantic,lin2016efficient,krahenbuhl2011efficient,li2021semantic}.
and
transformer structure~\cite{vaswani2017attention} based segmentation models~\cite{chen2021transunet,strudel2021segmenter,chen2021transunet,ranftl2021vision,strudel2021segmenter,zheng2021rethinking,xie2021segformer}.
Although significant performance is achieved, we
observe that the existing transformer based semantic segmentation models~\cite{cheng2021perpixel,cheng2022masked} are vulnerable to adversarial attack, making it necessary to analysis model robustness \wrt~adversarial attack for safer model deployment. In Fig.~\ref{fig:segmentation_robustness_wrt_transformer}, we illustrate performance of a transformer based segmentation model, namely Maskformer~\cite{cheng2021perpixel}, \wrt~adversarial attack, where the attack is generated by PGD~\cite{PGD_attack}. Fig.~\ref{fig:segmentation_robustness_wrt_transformer} shows that although Maskformer~\cite{cheng2021perpixel} produce accurate segmentation, it's not robust to adversarial attack, further strengthening our motivation in studying adversarial attacks. 

\noindent\textbf{Adversarial Attack for Semantic Segmentation:} is designed to evaluate robustness of semantic segmentation models with respect to adversarial attacks. Specifically,
\cite{arnab2018robustness} explores the adversarial robustness of the segmentation model using FGSM~\cite{Explaining_and_Harnessing_Adversarial_Examples} and PGD~\cite{PGD_attack}.
\cite{hendrik2017universal,fischer2017adversarial} generate universal perturbations for the semantic segmentation models and achieve targeted attack on the pedestrians. Note that universal attack is image-agnostic~\cite{moosavi2017universal,hendrik2017universal}, which is a small image perturbation that fools a deep neural network classifier to produce wrong predictions for almost all testing images.
Alternatively, targeted attack~\cite{adversarial_example_semantic_seg_iclr2017workshop,fischer2017adversarial,hendrik2017universal,li2021hidden} is a more fine-grained attack, where the attack is designed to fool the model to classify the image as a specific target class. 
%
For semantic segmentation, untargeted attacks are explored~\cite{agnihotri2023cospgd,gu2022segpgd}, where
SegPGD~\cite{gu2022segpgd} is presented as an adaptation of PGD~\cite{PGD_attack} for semantic segmentation, where misclassified pixels are identified to generate adversarial examples with fewer attack iterations. Further, \cite{gu2022segpgd,xu2021dynamic,xiao2018characterizing} use adversarial training in their framework to boost the robustness of semantic segmentation models, which focus on defense techniques instead of adversarial attacks.

\noindent\textbf{Transferable Adversarial Attacks:}~\cite{JiadongLin2019NesterovAG,zhu2022rethinking, CihangXie2018ImprovingTO, YinpengDong2019EvadingDT} are attacks that can transfer across different models. Conventionally, adversarial attacks are generated from a source model to attack the source model, which shows limitations in attacking the target models (see Fig.~\ref{semantic_seg_towards_attack}). Transferable attacks are generated from the source model, and they are required to be adversarial examples of the target model(s).
\cite{JiadongLin2019NesterovAG} uses the Nesterov accelerated gradient to replace the momentum method used in the previous gradient stabilized method~\cite{dong2018boosting} to optimize the gradient, leading to transferable attack for image classification.
~\cite{zhu2022rethinking} observes that samples in low-density regions correspond to adversarial examples with high transferability. They then define intrinsic attack as direction for the low-density region and introduce AAI (Alignment between Adversarial attack and Intrinsic attack) as an objective to identify samples with high transferability.
\cite{CihangXie2018ImprovingTO} augments the target image by resizing the image to a random size and then randomly padding zeros around the input image
to increase the transferability of the attack.
\cite{YinpengDong2019EvadingDT} applies three different kernel methods after getting the gradients, to make the perturbations more robust to the discriminative region of the model.

\noindent\textbf{Transferable Attack for Semantic Segmentation:}
DAG~\cite{xie2017adversarial} studies the transferability of attack for semantic segmentation and object detection via an adversary generation strategy, where the models are encouraged to produce a randomly generated class other than the target class. We find that DAG~\cite{xie2017adversarial} transfers well only on the cross-training setting, where the same network is trained with a different training dataset. When it comes to the different network structures, the transferability of DAG~\cite{xie2017adversarial} is poor, which is also consistent with our experiments. 
\cite{gu2021adversarial} investigates the overfitting phenomenon of adversarial examples for segmentation models, and attributes the transferability issue of semantic segmentation to the architectural traits of segmentation models, \ie~multi-scale object recognition. Dubbed dynamic scaling is then presented in \cite{gu2021adversarial} to achieve a transferable attack for semantic segmentation. Different from \cite{gu2021adversarial} that study transferability \wrt~model structures, we investigate transferable and conventional adversarial attacks for classification, and extend them to semantic segmentation to explain their transferability, aiming to provide algorithm-level understanding on the transferable attack for the dense prediction task.

We comprehensively investigate transferable adversarial attack for semantic segmentation (Sec.~\ref{sec_transferable_semantic_attack}). Different from the existing solutions via adversarial training~\cite{xie2017adversarial} or data scaling~\cite{gu2021adversarial}, we adapt the transferable attack for classification~\cite{JiadongLin2019NesterovAG,zhu2022rethinking, CihangXie2018ImprovingTO, YinpengDong2019EvadingDT} to semantic segmentation with a boarder view of the transferable semantic adversarial attack.
We first explain the existing adversarial attacks for semantic segmentation in Sec.~\ref{subsec_adveraial_attack_semantic}. Given the limited work on adversarial attack for semantic segmentation, we
extend the image classification based attacks (including both adversarial attacks and transferable attacks) to semantic segmentation in Sec.~\ref{adapting_existing_attack_for_segmentation}. We then analysis those transferable attacks in Sec.~\ref{analysis_attacks}, leading to the proposed
ensemble attack in Sec.~\ref{sec:ensemble_attack}, which is proven superior than the existing techniques in producing more transferable attacks for semantic segmentation.

\section{Adversarial Attacks for Semantic Segmentation}
\label{sec_transferable_semantic_attack}



As discussed previously, FGSM~\cite{Explaining_and_Harnessing_Adversarial_Examples} and PGD~\cite{PGD_attack} are two widely studied adversarial attacks for image classification. Due to its single-step nature, FGSM~\cite{Explaining_and_Harnessing_Adversarial_Examples} (see Eq.~\ref{eq_fgsm_attack})
has been demonstrated to be,
in general, less effective than its iterative version, namely PGD~\cite{PGD_attack} (see Eq.~\ref{eq_pgd_attack}). Although PGD~\cite{gu2022segpgd} is effective in image classification, \cite{gu2022segpgd} found that directly applying PGD~\cite{gu2022segpgd} to semantic segmentation can be less effective, as a large number of attack iterations are needed to generate accurate adversarial attack. \cite{gu2022segpgd} then presented SegPGD by identifying the misclassified pixels.


\subsection{Existing Adversarial Attack for Semantic Segmentation}
\label{subsec_adveraial_attack_semantic}
\noindent\textit{\textbf{SegPGD:}}
Specifically, \cite{gu2022segpgd} attributes the larger attack iterations of PGD~\cite{PGD_attack} for semantic segmentation to the imbalanced gradient contribution, where the wrongly classified pixels dominate the adversarial sample generation process, thus larger attack iterations are needed to mislead the correctly classified pixels with small cross-entropy loss. Based on the above observation, \cite{gu2022segpgd} divides the pixels into two groups according to the accuracy of the prediction, namely the accurately classified pixels $P^T$, and the wrongly classified ones $P^F$, leading to a reformulated loss function:
\begin{equation}
\mathcal{L}(\theta,\mathbf{x},\mathbf{y})=\frac{1}{H \times W}\sum_{j\in P^T}\mathcal{L}_i+\frac{1}{H \times W}\sum_{k\in P^F}\mathcal{L}_k,
\end{equation}
where $H$ and $W$ indicate the spatial dimension of the input,
$\mathcal{L}_i$ is the loss for the $i^{th}$ pixel, which is cross-entropy loss for semantic segmentation, and $\theta$ represents model parameters. 

To tackle the gradient dominant issue, \cite{gu2022segpgd} introduces dynamic weighted loss function:
\begin{equation}
\label{eq_segpgd}
\mathcal{L}(\theta,\mathbf{x}_{adv}^t,\mathbf{y})=\frac{1-\lambda_t }{H \times W}\sum_{j\in P^T}\mathcal{L}_i+\frac{\lambda_t}{H \times W}\sum_{k\in P^F}\mathcal{L}_k,
\end{equation}
where $\lambda_t$ is set dynamically with the number of attack iterations, and $\mathcal{L}_i=\mathcal{L}_i(\theta,\mathbf{x}_{adv}^t,\mathbf{y})$ is the loss for the $i^{th}$ pixel. Based on the reformulated loss function in Eq.~\ref{eq_segpgd}, the generated adversarial example with PGD~\cite{PGD_attack} can be achieved:
\begin{equation}
    \mathbf{x}_{adv}^{t+1} = \text{CLIP}(\mathbf{x}_{adv}^t + \alpha\text{sign} \left( \nabla_{\mathbf{x}_{adv}^t} \mathcal{L}_w(\theta, \mathbf{x}_{adv}^t, \mathbf{y}) \right),
\end{equation}
where $\nabla_{\mathbf{x}_{adv}^t} \mathcal{L}_w(\theta, \mathbf{x}_{adv}^t, \mathbf{y})$ is the weighted gradient with:
\begin{equation}
\begin{aligned} 
    &\mathcal{L}_w(\theta, \mathbf{x}_{adv}^t, \mathbf{y})\\
    &= \sum_{j\in P^T}(1-\lambda_t)\mathcal{L}_j(\theta, \mathbf{x}_{adv}^t, \mathbf{y}) + \sum_{j\in P^F}\lambda_t \mathcal{L}_k(\theta, \mathbf{x}_{adv}^t, \mathbf{y}),
\end{aligned}
\end{equation}
where $\alpha$ is the step size as in Eq.~\ref{eq_pgd_attack}. The initial point is usually defined as the clean sample, \ie~$\mathbf{x}_{adv}^0=\mathbf{x}_{clean}$, or its perturbed sample with uniform noise as $\mathbf{x}_{adv}^0=\mathbf{x}_{clean}+\mathcal{U}(-\epsilon,\epsilon)$ specified by the perturbation rate $\epsilon$. With the dynamic weighted loss function in Eq.~\ref{eq_segpgd}, SegPGD~\cite{gu2022segpgd} generates more effective adversarial examples than PGD~\cite{PGD_attack} under the same number of attack iterations.

\noindent\textit{\textbf{DAG:}} Although SegPGD~\cite{gu2022segpgd} takes wrongly classified pixels into consideration to speed up the adversarial sample generation process, there are no constraints on the extent to which the prediction should be destroyed. DAG~\cite{xie2017adversarial} can then be used to address the above issue, which is
designed to generate an adversarial perturbation that can confuse as many targets as possible.
Given one pixel $\mathbf{x}_{uv}$ in image $\mathbf{x}$ with coordinate $(u,v)$, with the pre-trained model $f(\theta)$, we obtain its classification score before softmax as $f(\mathbf{x}_{uv},\theta)\in\mathbb{R}^C$, where $C$ is the number of classes. DAG~\cite{xie2017adversarial} aims to cause all the pixel predictions to be incorrect.
With this goal, they define an objective:
\begin{equation}
\label{eq_dag_min_obj}
    \mathcal{L}=\sum_u\sum_v\left(f_{l_{uv}}(\mathbf{x}_{uv},\theta)-f_{l'_{uv}}(\mathbf{x}_{uv},\theta)\right),
\end{equation}
where $l_{uv}$ is the ground-truth label of $\mathbf{x}_{uv}$, and $l'_{uv}\in\{1,2,...,C\} \setminus \{l_{uv}\}$, which is other category than the actual one. Minimizing Eq.~\ref{eq_dag_min_obj} can cause every pixel to be incorrectly predicted. DAG~\cite{xie2017adversarial} then defines adversarial attack based on Eq.~\ref{eq_dag_min_obj} iteratively. For the $t^{th}$ iteration, they find the correctly predicted pixels, which can be denoted as $\mathbf{A}^t$. They obtain the accumulated perturbation as:
\begin{equation}
\label{eq_dag_accumulated_rt}
    \mathbf{r}^t = \sum_{uv\in \mathbf{A}^t}\left(\nabla_{\mathbf{x}^t}f_{l'_{uv}}(\mathbf{x}_{uv},\theta)-\nabla_{\mathbf{x}^t}f_{l_{uv}}(\mathbf{x}_{uv},\theta)\right).
\end{equation}
To achieve stable training, $\mathbf{r}^t$ is normalized as: $\mathbf{r'}^t=\frac{\gamma}{\|\mathbf{r}^t\|_\infty}\cdot \mathbf{r}^t$, where $\gamma=0.5$ in \cite{xie2017adversarial}. $\mathbf{x}^{t+1}$ is then obtained via $\mathbf{x}^{t+1}=\mathbf{r'}^t+\mathbf{x}^t$. The algorithm terminates when $\mathbf{A}^t=\varnothing$ or the maximum iteration is reached. The final perturbation is defined as: $\delta_{adv}=\sum_t \mathbf{r'}^t$, and the adversarial example is then $\mathbf{x}_{adv}=\mathbf{x}+\delta_{adv}$.

\subsection{Adapting Adversarial Attacks for Semantic Segmentation}
\label{adapting_existing_attack_for_segmentation}
Taking a step further, we extensively investigate transferable attack for semantic segmentation by extending the existing image classification based transferable attack to semantic segmentation. Four main strategies will be investigated, including Nesterov Iterative
Fast Gradient Sign Method (NI)~\cite{JiadongLin2019NesterovAG}, Diverse Inputs Iterative
Fast Gradient Sign Method (DI)~\cite{CihangXie2018ImprovingTO}, Translation-Invariant Attack Method (TI)~\cite{YinpengDong2019EvadingDT} and Intrinsic Adversarial Attack (IAA)~\cite{zhu2022rethinking}.


\noindent{\textit{\textbf{NI}
}}~\cite{JiadongLin2019NesterovAG} regards adversarial example generation as an optimization process
by adapting Nesterov accelarated gradient (NAG)~\cite{nesterov_accelerated_gradient} into the iterative attacks.
NAG~\cite{nesterov_accelerated_gradient} is a variation of normal gradient descent to speed up the training process with improved model convergence. Given dynamic variable $\mathbf{v}$, model parameters $\theta$, and loss function $\mathcal{L}$, NAG optimizes $\theta$ via:
\begin{equation}
    \begin{aligned}
\mathbf{v}^{t+1}=\mu\mathbf{v}^t+\nabla_{\theta_t}\mathcal{L}(\theta_t-\alpha\mu\mathbf{v}^t)\\
    \theta^{t+1}=\theta^t-\alpha\mathbf{v}^{t+1},
    \end{aligned}
\end{equation}
where $\mu$ is the decay factor of $\mathbf{v}^t$, and $\alpha$ is the step size as in Eq.~\ref{eq_pgd_attack}. NAG can be viewed as an improved momentum method~\cite{momentum_method}, which
is proven effective in achieving stabilized adversarial direction updating for transferable attack. Based on this, \cite{JiadongLin2019NesterovAG} introduces NI-FGSM with start point $\mathbf{g}^0=0$, which is formulated as:
\begin{equation}
\label{eq_ni_adversarial_attack}
    \begin{aligned}
        &\mathbf{x}^{nes}= \mathbf{x}_{adv}^t+\alpha\mu \mathbf{g}^t\\
        &\mathbf{g}^{t+1}=\mu\mathbf{g}^t + \frac{\nabla_{\mathbf{x}^{nes}} \mathcal{L}\left (\theta,\mathbf{x}^{nes},  \mathbf{y}\right )}{\left \| \nabla_{\mathbf{x}^{nes}} \mathcal{L}\left (\theta,\mathbf{x}^{nes},  \mathbf{y}\right ) \right \|_{1}}\\
        &\mathbf{x}_{adv}^{t+1}=\text{CLIP}( \mathbf{x}_{adv}^t+\alpha \text{sign} ( \mathbf{g}^{t+1} ) ),
    \end{aligned}
\end{equation}
where
$\mathbf{g}^t$ is the accumulated gradient,
$\mu$ is the decay rate of $\mathbf{g}^t$, $\mathbf{x}_{adv}$ is the Nesterov accelerated result. 
Eq.~\ref{eq_ni_adversarial_attack} provides an iterative adversarial example generation strategy, which is proven effective in generating transferable attack.



\noindent{\textit{\textbf{DI}}}~\cite{CihangXie2018ImprovingTO} observes iterative attacks tend to overfit the specific network parameters, leading to poor transferability. Inspired by data augmentation techniques to prevent the network from overfitting, \cite{CihangXie2018ImprovingTO} proposes to improves the transferability of adversarial examples by creating diverse input patterns, where the adversarial examples are then generated based on the diverse transformations of the input image.
Specifically, \cite{CihangXie2018ImprovingTO} augments the input image by resizing it
to a random size and then padding zeros around it.
Given transformation function $T(\cdot)$, \cite{CihangXie2018ImprovingTO} generates adversarial examples via:
\begin{equation}
\label{eq_di_attack}
    \mathbf{x}_{adv}^{t+1} = \text{CLIP}(\mathbf{x}_{adv}^t + \alpha\text{sign} \left( \nabla_{\mathbf{x}_{adv}^t} \mathcal{L}(\theta, T(\mathbf{x}_{adv}^t,p), \mathbf{y}) \right)),
\end{equation}
where the transformation function $T(\cdot)$ performs image augmentation with transformation probability $p$, \ie~with probability $p$, data augmentation will be applied to $\mathbf{x}_{adv}^t$, and probability $1-p$, $T(\mathbf{x}_{adv}^t)$ remains $\mathbf{x}_{adv}^t$.


\noindent{\textit{\textbf{TI}}}~\cite{YinpengDong2019EvadingDT} is another augmentation-based transferable attack by optimizing a perturbation over an ensemble of translated images, which can also be interpreted as convolving the gradient for the untranslated image with a pre-defined kernel $\mathbf{W}$. As TI~\cite{YinpengDong2019EvadingDT} directly works on the gradient of the loss function, it can be integrated into existing gradient based attack methods, \eg~FGSM~\cite{Explaining_and_Harnessing_Adversarial_Examples}, PGD~\cite{PGD_attack}, \etc. The combination of TI~\cite{YinpengDong2019EvadingDT} and PGD~\cite{PGD_attack} has the following update rule:
\begin{equation}
\label{eq_ti_pgd_attack}
    \mathbf{x}_{adv}^{t+1} = \text{CLIP}(\mathbf{x}_{adv}^t + \alpha\text{sign} \left( \mathbf{W}\ast\nabla_{\mathbf{x}_{adv}^t} \mathcal{L}(\theta, \mathbf{x}_{adv}^t, \mathbf{y}) \right)).
\end{equation}
where the kernel matrix $\mathbf{W}$ is selected as a
Gaussian kernel, which is used to convolve
the gradient, namely $\nabla_{\mathbf{x}_{adv}^t}\mathcal{L}$.

\noindent{\textit{\textbf{IAA}}}~\cite{zhu2022rethinking}
investigates transferable attack from the data distribution perspective, and observe that samples in low density region of the data distribution have much stronger transferability than those in the high density region.
Based on the above observation, \cite{zhu2022rethinking} first introduce intrinsic attack as direction towards the low density region, namely $-\nabla_{\mathbf{x}} \log p(\mathbf{x}, \mathbf{y})$. Then a metric is introduced to evaluate the alignment of the model’s adversarial
attack with intrinsic attack, which is defined as \textbf{A}lignment
between \textbf{A}dversarial attack and the \textbf{I}ntrinsic attack (AAI):
\begin{equation}
\label{eq_aai_definition}
    \begin{aligned}
     \mathbf{\text{AAI}} \triangleq \mathbb{E}_{p(\mathbf{x}, \mathbf{y})}\left[\frac{\nabla_{\mathbf{x}} \log p_{\theta, \lambda}(\mathbf{y} \mid \mathbf{x})}{\left\|\nabla_{\boldsymbol{\mathbf{x}}} \log p_{\theta, \lambda}(\mathbf{y} \mid \mathbf{x})\right\|_2} \cdot \nabla_{\mathbf{x}} \log p(\mathbf{x}, \mathbf{y})\right],
    \end{aligned}
\end{equation}
where $p(\mathbf{x}, \mathbf{y})$ is the true data joint distribution, and $p_{\theta, \lambda}(\mathbf{y} \mid \mathbf{x})$ is the conditional generation model with $\theta$ as the pre-trained conditional generation model $p_{\theta}(\mathbf{y} \mid \mathbf{x})$, and $\lambda$ represents the structure level modification of $p_{\theta}(\mathbf{y} \mid \mathbf{x})$ towards transferable attack, which include replacing ReLU~\cite{agarap2018deep} with SoftPlus~\cite{zheng2015improving} and updating residual connections~\cite{he2016deep} to perparameterized residual connections. 
$\lambda$ in Eq.~\ref{eq_aai_definition} can be learned via Bayesian optimization. Particularly, as unobserved data distribution $p(\mathbf{x}, \mathbf{y})$ is involved in the optimization function, \cite{zhu2022rethinking} uses score matching~\cite{https://doi.org/10.48550/arxiv.1905.07088} together with Bayesian optimization to
search for $\lambda$.

\subsection{Criteria for Transferable Attack}
\label{analysis_attacks}
To achieve transferable attack, 
\cite{dong2018boosting} believes that generalization capacity of the trained models should be considered.
Generating adversarial examples is analogous to optimizing neural networks during training. In this process, the white-box model under attack is refined using adversarial examples, which serve as training data. These examples can be regarded as the training parameters for the model. During testing, black-box models evaluate the adversarial examples, acting as testing data. Consequently, techniques employed to improve model generalization can be applied to enhance the transferability of adversarial examples. 


The generalization process has two main ways, namely
data augmentation and
better optimization. For former, Diverse Inputs Iterative
Fast Gradient Sign Method (DI)
\cite{CihangXie2018ImprovingTO} and Translation-Invariant Attack Method (TI)~\cite{YinpengDong2019EvadingDT} believe that transfer to an unseen model should be similar to transfer to some unseen images. In this way,
they try to augment the images before inputting them into the attack technique, \eg~PGD~\cite{PGD_attack}.
For the latter,
the NI~\cite{JiadongLin2019NesterovAG} use a Nesterov Accelerated Gradient \cite{nesterov2012efficiency} in the gradient descent stage, which applies the idea of momentum. Fast Gradient Sign Method (NI)~\cite{JiadongLin2019NesterovAG} proves that although different networks will have different decision boundaries (due to their high non-linearity), they still have similar test performance \cite{liu2016delving}. As a surrogate refinement method, the Intrinsic Adversarial Attack (IAA)~\cite{zhu2022rethinking} defines AAI to indicate the correlation between the global optimize direction and the current direction. It changes the structure parameters to get the maximum AAI, achieving perfect alignment of intrinsic attack with the model's current adversarial attack for high transferabibility.

Based on above analysis, to realize high transferability, we conclude that an attack method has to be attacked on the critical region as IAA~\cite{zhu2022rethinking}, or augmented to deal with unseen models, or use a stabilized gradient to try to find a global optimal attack direction, leading to our ensemble attack.  We aim to aggregate multiple transferable attacks for classification to extensively explore their contributions for transferable attack.

\section{Ensemble attack}
\label{sec:ensemble_attack}
Given the different transferability of existing attacks for semantic segmentation (see Fig.~\ref{fig:transferability_evaluation}), we intend to
extensively explore their potential for transferable attack, which is also named \enquote{ensemble segmentation attack}. Different from~\cite{cai2023ensemblebased}, which introduces an ensemble of surrogate models, where weights of the involved models are optimized based on the victim model, we work on attack-ensemble, and generate attack based on an ensemble of different segmentation attacks. 
Specifically, given the superiority of NI~\cite{JiadongLin2019NesterovAG},
DI~\cite{CihangXie2018ImprovingTO} and
TI~\cite{YinpengDong2019EvadingDT} for transferable attack with different focus as discussed in Sec.~\ref{analysis_attacks}, we aggregate adversarial attacks from them
achieving ensemble semantic attack.
To achieve ensemble attack, we apply DI~\cite{CihangXie2018ImprovingTO} and
TI~\cite{YinpengDong2019EvadingDT} on the data to use more augmentation of the data and to utilize those invariant features. Then, we rely on NI~\cite{JiadongLin2019NesterovAG} to
incorporate
the accelerated gradient of the above to avoid the attack coming into a local minimal, leading to our ensemble attack as: 
\begin{equation}
\label{eq_ni_di_ti_adversarial_attack}
    \begin{aligned}
        &\mathbf{x}^{nes}= \mathbf{x}_{adv}^t+\alpha\mu \mathbf{g}^t\\
        &\mathbf{g}^{t+1}=\mu\mathbf{g}^t + \frac{\mathbf{W} * \nabla_{\mathbf{x}^{nes}} \mathcal{L}(\theta, T(\mathbf{x}^{nes},p), \mathbf{y}) }{\left \|  \mathbf{W} * \nabla_{\mathbf{x}^{nes}} \mathcal{L}(\theta, T(\mathbf{x}^{nes},p), \mathbf{y}) \right \|_{1}}\\
        &\mathbf{x}_{adv}^{t+1}=\text{CLIP}( \mathbf{x}_{adv}^t+\alpha \text{sign} ( \mathbf{g}^{t+1} ) ),
    \end{aligned}
\end{equation}
where the transformation function $T(\cdot)$ performs image augmentation with transformation probability $p$,
the kernel matrix $\mathbf{W}$ is selected as a
Gaussian kernel, and $\mathbf{g}^t$ is the accumulated gradient,
$\mu$ is the decay rate of $\mathbf{g}^t$, $\mathbf{x}^{nes}$ is the Nesterov accelerated result.
The experimental results show that the ensemble method performs better than the strongest single method (NI~\cite{JiadongLin2019NesterovAG}), leading to better transferability.

\section{Experiments}
We investigate adversarial attack for semantic segmentation and explain their transferability. As there exists no benchmark setting for semantic segmentation, we re-implement all the related models in this section with the same setting for a fair comparison.
\begin{table*}[t!]
  \centering
  \footnotesize
  \renewcommand{\arraystretch}{1.2}
  \renewcommand{\tabcolsep}{4.2mm}
  \caption{Adversarial attack on PascalVOC dataset, where attacks with each individual source model, e.g. \enquote{DV3Res50}, are categorised as the
  conventional attacks (those above the middle line) and the
  transferable attacks (those below the middle line). The bold numbers highlight the best performance based on each source model.
  }
  \begin{tabular}{l|c|cc|cc|cc|cc}
  \toprule[1.1pt]
  &&\multicolumn{2}{c|}{Image Quality}&\multicolumn{6}{c}{Target (performance is evaluated with $\text{mIOU}\downarrow$ (left) and $\text{Sr}\uparrow$ (right))}\\ \hline
  Source & Attack & $\text{PSNR}\uparrow$ & $\text{SSIM}\uparrow$ & \multicolumn{2}{c|}{DV3Res50} & \multicolumn{2}{c|}{DV3Res101}& \multicolumn{2}{c}{FCNVGG16}  \\ \hline
  \multirow{9}{*}{DV3Res50}  & FGSM~\cite{Explaining_and_Harnessing_Adversarial_Examples} & 30.18 & 0.5098 & 0.3400 & 0.5480 & 0.5134 & 0.3410 &  0.4628 & 0.2890 \\ 
  & PGD~\cite{PGD_attack} & 39.70 & 0.7814 & 0.0938 & 0.8753 & 0.4232 & 0.4568 &  0.5681 & 0.1273   \\ 
  & SegPGD~\cite{gu2022segpgd} & 40.13 & 0.7897 & 0.1127 & 0.8502 & 0.5659 & 0.2736 &   0.6030 & 0.0740 \\ 
   & DAG~\cite{xie2017adversarial} & \textbf{46.09} & \textbf{0.9340} & 0.1268 & 0.8314 & 0.6271 & 0.1951 & 0.5977 & 0.0819 \\ 
   \cline{2-10}
      &  NI~\cite{JiadongLin2019NesterovAG} & 32.84 & 0.5912 & \textbf{0.0861} & \textbf{0.8855} & \textbf{0.2610}  &\textbf{0.6650} &  \textbf{0.4246} & \textbf{0.3478}  \\ 
  & DI ~\cite{CihangXie2018ImprovingTO} & 39.70 & 0.7793 & 0.0950 & 0.8737 &  0.2929 & 0.6241 &  0.5035 & 0.2266  \\ 
  & TI ~\cite{YinpengDong2019EvadingDT} & 39.26 & 0.7808 & 0.0997 & 0.8675 & 0.3871 & 0.5031 & 0.5193 & 0.2023 \\
    & IAA ~\cite{zhu2022rethinking} & 35.69 & 0.9396 & 0.0823 & 0.8897 & 0.5379 & 0.1302 & 0.5625 & 0.1360  \\
   \midrule[0.7pt]
   \multirow{9}{*}{DV3Res101} & FGSM~\cite{Explaining_and_Harnessing_Adversarial_Examples} & 30.17  & 0.5104 & 0.4959 & 0.3407 & 0.3774 & 0.5156 &  0.4764 & 0.2682 \\ 
   & PGD~\cite{PGD_attack} &  39.64  & 0.7804 & 0.3564 & 0.5262 & 0.1001  & 0.8715 & 0.5673 & 0.1286 \\ 
  & SegPGD~\cite{gu2022segpgd} & 40.09 & 0.7891 & 0.4557 & 0.3941 & 0.1131 & 0.8548 & 0.5985 & 0.0806  \\
   & DAG~\cite{xie2017adversarial} &\textbf{46.41} & \textbf{0.9401} & 0.5710 & 0.2409 & 0.1569 & 0.7986 & 0.6057 & 0.0696  \\ 
   \cline{2-10}
    &  NI~\cite{JiadongLin2019NesterovAG} & 32.83 & 0.5925 & \textbf{0.2244} & \textbf{0.7017} & \textbf{0.0922} & \textbf{0.8817} & \textbf{0.4341}  & \textbf{0.3332 } \\ 
    & DI ~\cite{CihangXie2018ImprovingTO} & 39.70 & 0.7791 & 0.2545 & 0.6617 & 0.0973 & 0.8751 &  0.5029 & 0.2275  \\ 
    & TI ~\cite{YinpengDong2019EvadingDT} & 39.26 & 0.7805 & 0.3543 & 0.5290 & 0.1068  & 0.8629 & 0.5235& 0.1959  \\
  
    & IAA ~\cite{zhu2022rethinking} & 34.21& 0.9244 & 0.4597 & 0.3481 & 0.1315 & 0.7874  & 0.5690 & 0.1259 \\
\midrule[0.7pt]
  \multirow{8}{*}{FCNVGG16} & FGSM~\cite{Explaining_and_Harnessing_Adversarial_Examples} & 30.18 & 0.5099 & 0.3811 & 0.4933 & 0.4641 & 0.4043 &  0.1715 & 0.7366  \\ 
  & PGD~\cite{PGD_attack} & 36.73 & 0.5106  & 0.1989 & 0.7356 &  0.2981  & 0.6174  & \textbf{0.0309} & \textbf{0.9525}  \\ 
  & SegPGD~\cite{gu2022segpgd} & 37.86 & 0.6023 & 0.3346 & 0.5552 &  0.4337 & 0.4433 & 0.0608 & 0.9066  \\
   & DAG~\cite{xie2017adversarial}  &\textbf{42.64} & \textbf{0.9216} & 0.4723 & 0.3721 & 0.5538 & 0.2892 & 0.1378 & 0.7883 \\
    \cline{2-10}
   &  NI~\cite{JiadongLin2019NesterovAG} & 32.75 & 0.6018 & \textbf{0.1607} & \textbf{0.7864} & \textbf{0.2233} & \textbf{0.7134} &  0.0326 & 0.9499  \\ 
    & DI ~\cite{CihangXie2018ImprovingTO} & 37.39 & 0.7492 & 0.1747 & 0.7677 & 0.2536  & 0.6745 &  0.0335 & 0.9485  \\ 
    & TI ~\cite{YinpengDong2019EvadingDT} & 36.35 & 0.7382 & 0.2272 & 0.6384 & 0.3262  & 0.5813 & 0.0320 & 0.9508 \\
\bottomrule[1.1pt]
  \end{tabular}
  \label{tab:pascalvoc_adversarial_attack}
\end{table*}

\subsection{Setting}
\noindent\textbf{Dataset:}
We train our models on the augmented Pascal VOC 2012~\cite{pascal-voc-2012}
dataset, which contains 21 classes with 10,582 images. 
During training, we also perform online data augmentation by applying a random scaling with ratio in the range of $[0.5, 2]$,
a random crop with the size of 513x513, and a random horizontal flip. 
As semantic segmentation models can not generalize to datasets with different categories, we also train our models on the Cityscape~\cite{Cordts2016Cityscapes} training dataset, which contains 3,475 images with 19 classes.
During the training stage with Cityscape dataset, 
we apply random crop with size of 768x768, color jitter, and random horizontal flip for data augmentation. 
We evaluate the related models on their corresponding testing datasets of size 1,449 and 1,525 for Pascal VOC 2012~\cite{pascal-voc-2012} and Cityscape~\cite{Cordts2016Cityscapes}, respectively, where the spatial size of the former is 513x513 for testing, and we use the raw testing images without performing spatial scaling for the latter, following the conventional setting.

\noindent\textbf{Measures:}
We evaluate model accuracy for semantic segmentation with mean IoU ($\text{mIoU}$).
To model the transferability of adversarial attacks, we first use Peak Signal-to-Noise Ratio ($\text{PSNR}$) and
Structural Similarity Index Measure (SSIM)~\cite{wang2004image}
to measure the noisy degree of the adversarial example, as the adversarial examples should be $L_p$-normed, indicating a visually invisible perturbation. Then, we adopt success rate ($\text{Sr}$) to measure the effectiveness of the adversarial attack, which is defined as the proportion of successfully destroyed model prediction as: $S_r = 1 - \frac{\text{mIoU}_\text{adv}}{\text{mIoU}_\text{clean}}$, where $\text{mIoU}_\text{adv}$ and $\text{mIoU}_\text{clean}$ are the segmentation performance of the adversarial example $\mathbf{x}_\text{adv}$ and the clean image $\mathbf{x}$, respectively.

\noindent\textbf{Adversarial Attacks:} We first investigate the conventional adversarial attacks, including
FGSM~\cite{Explaining_and_Harnessing_Adversarial_Examples},
PGD~\cite{PGD_attack}, and apply them to semantic segmentation. We also analyse the transferability of existing semantic attacks, namely SegPGD~\cite{gu2022segpgd} and DAG~\cite{xie2017adversarial}. 
Further, we adapt the existing transferable attack for image classification to our dense semantic segmentation task, including NI~\cite{JiadongLin2019NesterovAG},
DI~\cite{CihangXie2018ImprovingTO},
TI~\cite{YinpengDong2019EvadingDT}, and AAI~\cite{zhu2022rethinking}.

\noindent\textbf{Network structures:}
We investigate different backbones for adversarial attack transferability analysis.
Specifically, for the PascalVOC dataset, we investigate Deeplabv3~\cite{chen2017rethinking} with ResNet50 backbone (\enquote{DV3Res50}), Deeplabv3+~\cite{chen2018encoder} with ResNet101 backbone (\enquote{DV3Res101}) and FCN~\cite{long2015fully} with VGG16 backbone (\enquote{FCNVGG16}). For Cityscape dataset, we run models using Deeplabv3~\cite{chen2017rethinking} (\enquote{DV3Res50}) with ResNet50 backbone, Deeplabv3+~\cite{chen2018encoder} (\enquote{DV3Res101}) with ResNet101 backbone and Deeplabv3+~\cite{chen2017rethinking} with MobilenetsV2~\cite{sandler2018mobilenetv2} backbone (\enquote{DV3MOB}). We show parameter numbers of each network structure in Table~\ref{tab:backbone_parameter_numbers} for a clear comparison of the sizes of the related models.
\begin{table*}[t!]
  \centering
  \footnotesize
  \renewcommand{\arraystretch}{1.2}
  \renewcommand{\tabcolsep}{4.2mm}
  \caption{Adversarial attack on Cityscape dataset, where attacks are grouped, and best performance numbers are highlighted.}
  \begin{tabular}{l|c|cc|cc|cc|cc}
  \toprule[1.1pt]
  &&\multicolumn{2}{c|}{Image Quality}&\multicolumn{6}{c}{Target (performance is evaluated with $\text{mIOU}\downarrow$ and $\text{Sr}\uparrow$)}\\ \hline
  Source & Attack & $\text{PSNR}\uparrow$ & $\text{SSIM}\uparrow$ & \multicolumn{2}{c|}{DV3Res50} & \multicolumn{2}{c|}{DV3101}& \multicolumn{2}{c}{DV3MOB} \\ \hline
  \multirow{7}{*}{DV3Res50} & FGSM~\cite{Explaining_and_Harnessing_Adversarial_Examples}  & 30.11 & 0.2870 & 0.2834 & 0.6343 &  0.3892 & 0.4903 & 0.2834 & 0.6106  \\ 
  & PGD~\cite{PGD_attack}  & 39.92 & 0.6066 & 0.0108 & 0.9861 & 0.2990  & 0.6084  &  0.3987 & 0.4522 \\
    
  & SegPGD~\cite{gu2022segpgd} & 40.34 & 0.6195 & 0.0210 & 0.9729 & 0.3550 & 0.5351 & 0.4771 & 0.3444  \\ 
    & DAG~\cite{xie2017adversarial} & \textbf{43.91}  & \textbf{0.7758} & 0.0761 & 0.9018 & 0.4895 & 0.3590 & 0.5476 & 0.2476  \\ 
    \cline{2-10}
    & NI~\cite{JiadongLin2019NesterovAG}  & 32.91 & 0.3729 & 0.0142 & 0.9817 & \textbf{0.1398}  & \textbf{0.8169} &  \textbf{0.1754} & \textbf{0.7590}  \\ 
    & DI ~\cite{CihangXie2018ImprovingTO}  & 39.90 & 0.6036 & 0.0178 & 0.9770 & 0.1925  & 0.7479 & 0.2580 & 0.6455  \\ 
    
   & TI ~\cite{YinpengDong2019EvadingDT} & 39.70 & 0.6119 & \textbf{0.0103} & \textbf{0.9867} & 0.2562  & 0.6645 & 0.3231 & 0.5561 \\ 
   \midrule[0.7pt]
   \multirow{8}{*}{DV3Res101} & FGSM~\cite{Explaining_and_Harnessing_Adversarial_Examples}  & 30.11 & 0.2890 & 0.3103 & 0.5996 &  0.3424  & 0.5516 &  0.2655 & 0.6352  \\ 
   & PGD~\cite{PGD_attack}  & 39.89 & 0.6064 & 0.2064  & 0.7337 & 0.0164 & 0.9785 & 0.3875 & 0.4676  \\ 
  & SegPGD~\cite{gu2022segpgd}  & 40.33 & 0.6204 & 0.2542 & 0.6720 &  0.0251 & 0.9671 &  0.4669 & 0.3585  \\
    & DAG~\cite{xie2017adversarial} & \textbf{44.39} & \textbf{0.8026} & 0.4412 & 0.4307 &  0.1151 & 0.8493 & 0.5560 & 0.2361  \\ 
    \cline{2-10}
    & NI~\cite{JiadongLin2019NesterovAG}  & 32.91 & 0.3736 & \textbf{0.1102} & \textbf{0.8578} & 0.0176 & 0.9770 &  \textbf{0.1665} & \textbf{0.7712}  \\ 
    & DI ~\cite{CihangXie2018ImprovingTO} & 39.88 & 0.6045 &  0.1544 & 0.8008 &  0.0268 & 0.9649 &  0.2683 & 0.6314 \\ 
       & TI ~\cite{YinpengDong2019EvadingDT}  & 39.69 & 0.6135 & 0.1984 & 0.7440 &  \textbf{0.0164} & \textbf{0.9785} & 0.3271 & 0.5506  \\
\midrule[0.7pt]
  \multirow{7}{*}{DV3MOB} & FGSM~\cite{Explaining_and_Harnessing_Adversarial_Examples}  & 30.12 & 0.2887 & 0.3356 & 0.5670 & 0.4019 & 0.4737  &  0.2072 & 0.7153 \\ 
  & PGD~\cite{PGD_attack}  & 39.97 & 0.6085 & 0.4743 & 0.3880 & 0.5450 & 0.2863 &  0.0110 & 0.9849 \\ 
  & SegPGD~\cite{gu2022segpgd}  & 40.40 & 0.6217 & 0.6077 & 0.2159 & 0.6745 & 0.1167 &  0.0221 & 0.9696 \\
    & DAG~\cite{xie2017adversarial} & \textbf{43.27} & \textbf{0.7420}& 0.5826 & 0.2483 & 0.6306 & 0.1742  & 0.0330 & 0.9547 \\ 
    \cline{2-10}
    & NI~\cite{JiadongLin2019NesterovAG}  & 32.81 & 0.3739 & \textbf{0.2976} & \textbf{0.6160} &  \textbf{0.3458} & \textbf{0.5471} &  0.0109 & 0.9850 \\ 
    & DI ~\cite{CihangXie2018ImprovingTO}  & 39.89 & 0.6035 & 0.3400 & 0.5613 & 0.4350  & 0.4303 &  0.0128 & 0.9824  \\ 
       & TI ~\cite{YinpengDong2019EvadingDT}  & 39.62 & 0.6137 & 0.3886 & 0.4986 & 0.4796 & 0.3719 & \textbf{0.0064} & \textbf{0.9912} \\
\bottomrule[1.1pt]
  \end{tabular}
  \label{tab:adversarial_cityscape}
\end{table*}
\noindent\textbf{Adversarial attack:} The attack rate $\epsilon$ is restricted to 0.03 (8/255 in $L_\infty$ norm) to ensure that the adversarial example has enough perturbation to fool the segmentation model, while it is still visually invisible.
For the iterative attacks (including DAG~\cite{xie2017adversarial}), we set the attack iteration as 10 to achieve a trade-off between
attack quality and model efficiency.
\begin{figure*}[tp]
   \begin{center}
   \begin{tabular}{{c@{ } c@{ } c@{ }c@{ } c@{ } }}
{\includegraphics[width=0.185\linewidth]{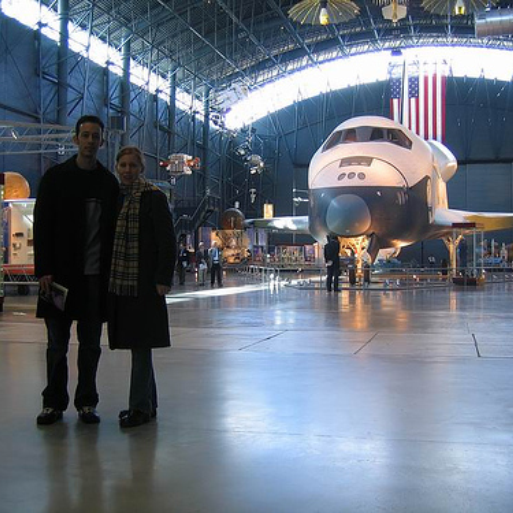}}&  
{\includegraphics[width=0.185\linewidth]{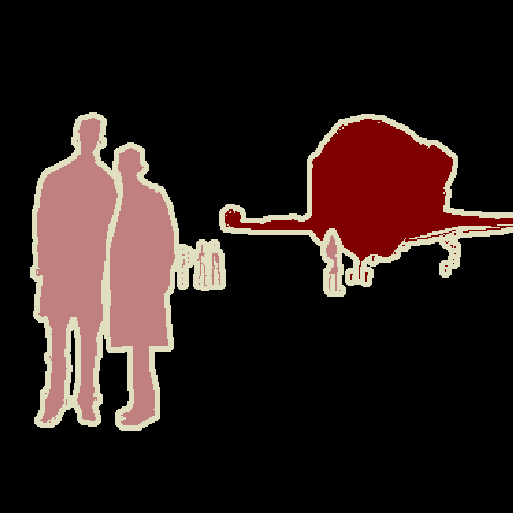}}&  
{\includegraphics[width=0.185\linewidth]{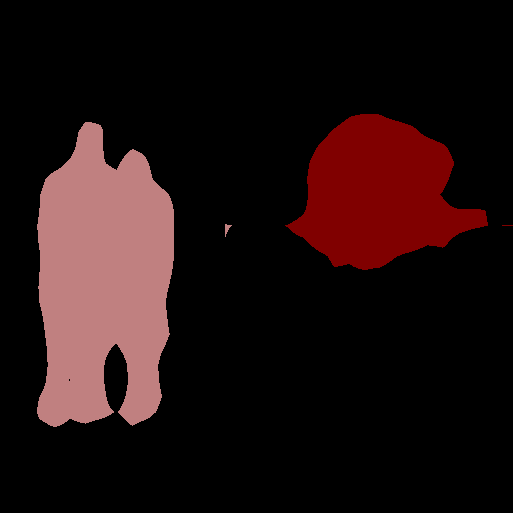}}&  
{\includegraphics[width=0.185\linewidth]{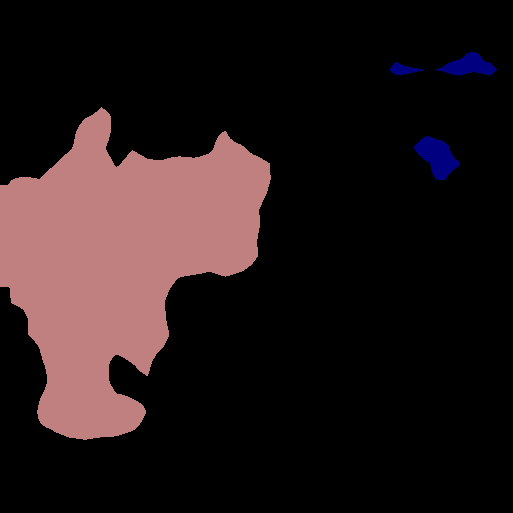}}&
{\includegraphics[width=0.185\linewidth]{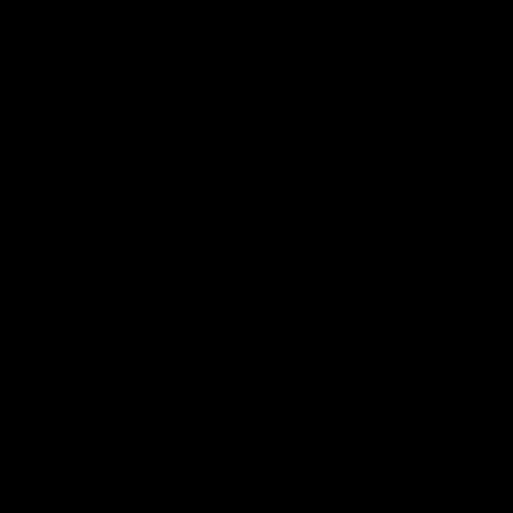}}\\ 
\footnotesize{Image}&\footnotesize{GT}&\footnotesize{Original}&\footnotesize{FGSM~\cite{Explaining_and_Harnessing_Adversarial_Examples}}&\footnotesize{SegPGD~\cite{gu2022segpgd}}\\
{\includegraphics[width=0.185\linewidth]{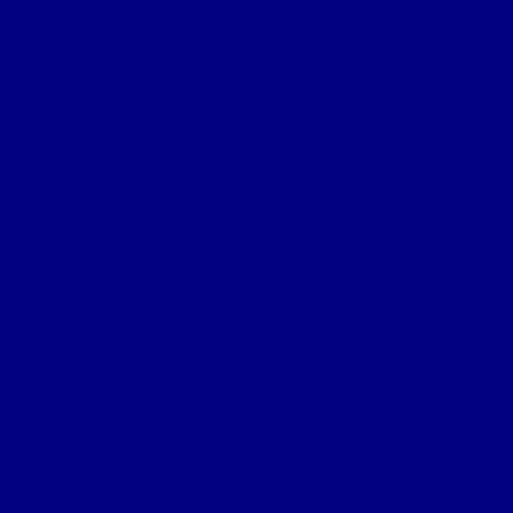}}&  
{\includegraphics[width=0.185\linewidth]{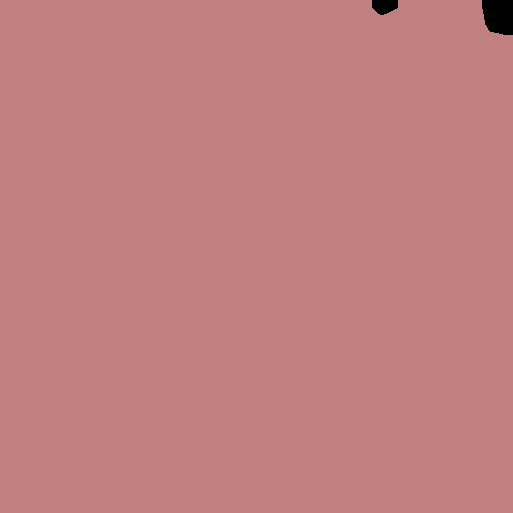}}&  
{\includegraphics[width=0.185\linewidth]{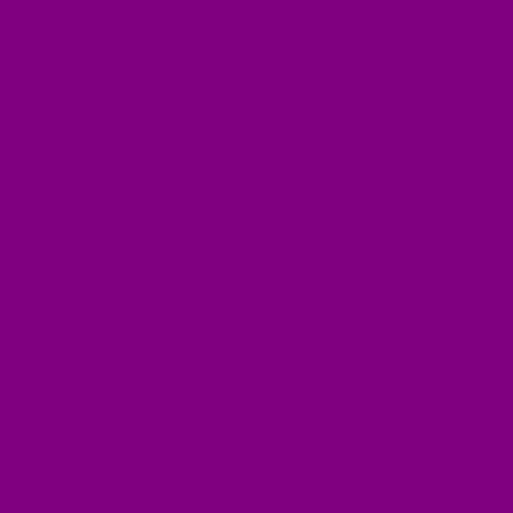}}&  
{\includegraphics[width=0.185\linewidth]{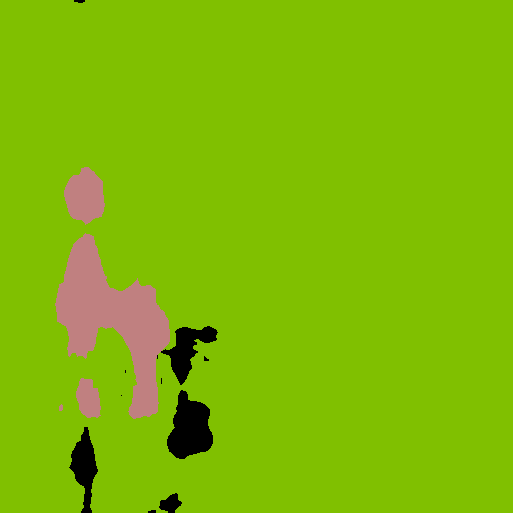}}&
{\includegraphics[width=0.185\linewidth]{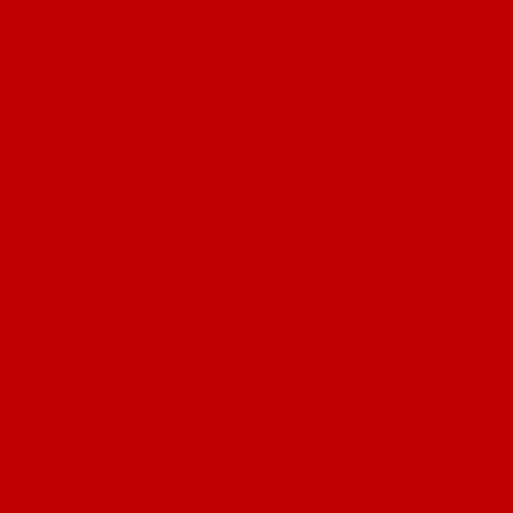}}\\ 
\footnotesize{PGD~\cite{PGD_attack}}&\footnotesize{DI~\cite{CihangXie2018ImprovingTO}}&\footnotesize{IAA~\cite{zhu2022rethinking}}&\footnotesize{DAG~\cite{xie2017adversarial}}&\footnotesize{NI~\cite{JiadongLin2019NesterovAG}}\\
   \end{tabular}
   \end{center}
    \caption{Predictions of adversarial examples generated with different attacks on PascalVOC dataset.
    } 
    \label{pascal_voc_attack_performance}
\end{figure*}

\subsection{Performance Comparison}
In Table~\ref{tab:pascalvoc_adversarial_attack} and Table~\ref{tab:adversarial_cityscape}, we show model performance of conventional and transferable attacks on the PascalVOC 2012 testing dataset and Cityscape dataset, respectively. We show both quality of the adversarial examples ($\text{PSNR}$ and $\text{SSIM}$), prediction accuracy of segmentation results ($\text{mIoU}$) as well as attack quality ($\text{S}_r$).
Both Table~\ref{tab:pascalvoc_adversarial_attack} and Table~\ref{tab:adversarial_cityscape} show that DAG~\cite{xie2017adversarial} produces attack with high image quality, and performs well on the source model, but fails to transfer to the target model, which is also consistent with the results in \cite{xie2017adversarial}. Among all the transferable attacks, namely NI~\cite{JiadongLin2019NesterovAG},
DI~\cite{CihangXie2018ImprovingTO},
TI~\cite{YinpengDong2019EvadingDT}, and IAA~\cite{zhu2022rethinking}, NI~\cite{JiadongLin2019NesterovAG} achieves overall the best transferability, indicating the importance of Nesterov Accelerated Gradient \cite{nesterov2012efficiency} with stabilized training for transferable attack. However, compared with DAG~\cite{xie2017adversarial}, we observe degraded image quality of NI~\cite{JiadongLin2019NesterovAG}, explaining the trade-off between transferability and strength of the input perturbation.
Another observation from Table~\ref{tab:pascalvoc_adversarial_attack} and Table~\ref{tab:adversarial_cityscape} is that the existing transferable attacks transfer well across to the residual-connection models, \ie~Deeplabv3 and Deeplabv3+, while we find inferior transferability cross to the residual and non-residual based network, \ie~Deeplabv3 and FCN with VGG16 backbone. More investigation will be conducted to analysis model structure \wrt~transferability of attacks. 

\begin{figure}[t!]
   \begin{center}
   \begin{tabular}{{c@{ }}}

{\includegraphics[width=0.95\linewidth]{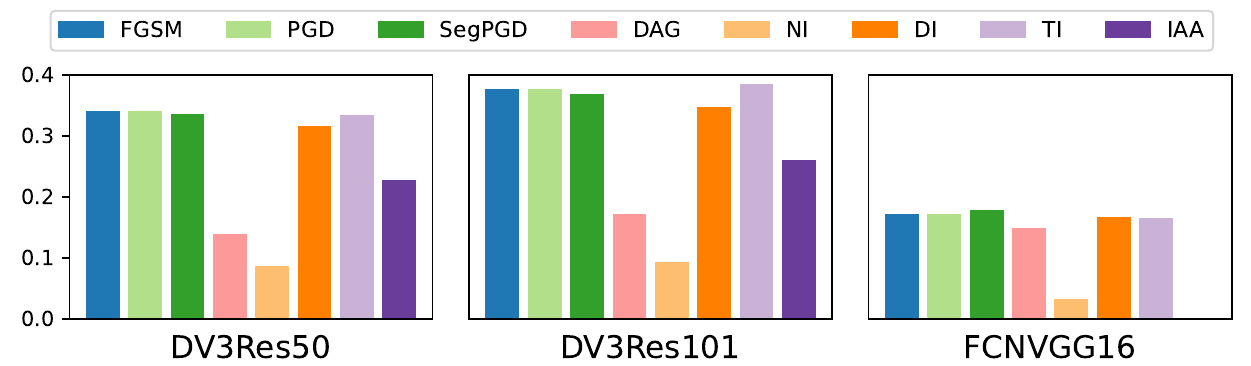}}\\
   \end{tabular}
   \end{center}
    \caption{Model mean IOU \wrt~adversarial attacks, where we generate adversarial attack based on each individual model to evaluate semantic segmentation model robustness \wrt~adversarial attack.} 
    \label{fig:model_performance_wrt_conventional_ad_attack}
\end{figure}

\noindent\textbf{Model Robustness \wrt~Adversarial Attack:} For each testing dataset, with a specific source model, we want to evaluate model robustness \wrt~different adversarial attacks. Taking PascalVOC dataset for example, we show the mean IOU of the testing dataset after applying each adversarial attack with three network structures in
Fig.~\ref{fig:model_performance_wrt_conventional_ad_attack}.
Note that, the source and target model, in this case, are the same. Ideally, we want the attack to be effective and the model should perform poorly for the adversarial example. Fig.~\ref{fig:model_performance_wrt_conventional_ad_attack} shows that, for all the three network structures,
NI~\cite{JiadongLin2019NesterovAG} achieves the best attack performance with the lowest $\text{mIoU}$, and DAG~\cite{xie2017adversarial} is the best in image quality. The result in Fig.~\ref{fig:model_performance_wrt_conventional_ad_attack} makes sense in that NI~\cite{JiadongLin2019NesterovAG} is an improved momentum method~\cite{momentum_method}, which is more stable to train, leading to more effective adversarial example. DAG~\cite{xie2017adversarial} introduces a new objective (see Eq.~\ref{eq_dag_min_obj}) to explicitly force the model to produce inaccurate predictions.

\begin{table}[t!]
    \caption{\#Model parameters (measured in millions).
    }
    \label{tab:backbone_parameter_numbers}
    \centering
    \footnotesize
    \setlength{\tabcolsep}{1.5mm}{
        \begin{threeparttable}
        \begin{tabular}{ccccc}
        \toprule[1.1pt]
         & DV3Res50    & DV3Res101 & FCNVGG16 & DV3MOB      \\
        \midrule
        \#Parameters    & 23.5 & 42.5 & 134.5 & 2.2  \\
        \bottomrule[1.1pt]
        \end{tabular}
        \end{threeparttable}
    }
\end{table}

\begin{figure*}[tp]
   \begin{center}
   \begin{tabular}{{c@{ }}}
{\includegraphics[width=0.90\textwidth]{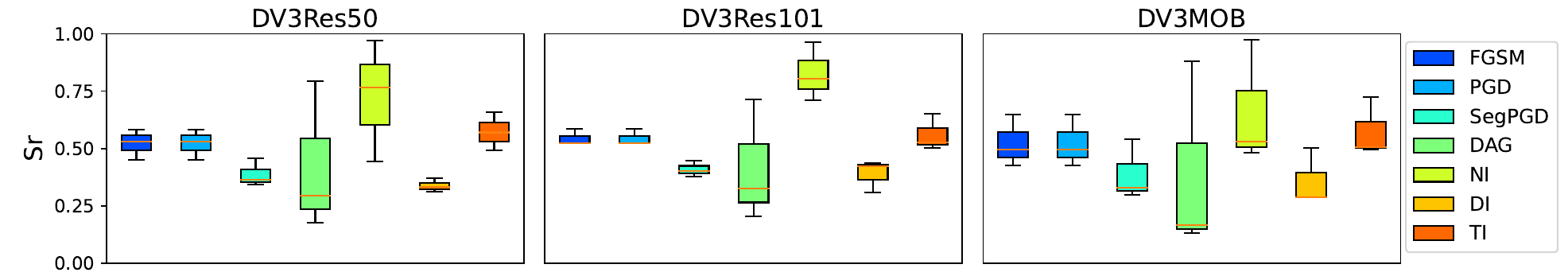}}\\
   \end{tabular}
   \end{center}
    \caption{Range of the MIoU after attacked (y-axis) of different attacks on the Cityscape dataset with different source models, where the MIoU is computed based on one specific source model and the other two target models, e.g. DV3Res50 as source model, while DV3Res101 and DV3MOB as target models.
    }
    \label{fig:transferability_evaluation}
\end{figure*}

We also show a visual comparison of different attacks in Fig.~\ref{pascal_voc_attack_performance}, which is based on the PascalVOC dataset with DV3Res50 backbone. The results show that the original prediction is reasonable compared with the ground truth. 
With FGSM~\cite{Explaining_and_Harnessing_Adversarial_Examples} attack, the foreground person can still be observed from the segmentation map. However, the plane has almost disappeared. IAA~\cite{zhu2022rethinking} is a distribution alignment-based transferable attack. In this source-to-source setting, its performance is inferior compared with DAG~\cite{xie2017adversarial} and NI~\cite{JiadongLin2019NesterovAG}, where in DAG~\cite{xie2017adversarial} the foreground is almost disappeared, and in NI~\cite{JiadongLin2019NesterovAG}, PGD~\cite{PGD_attack}, IAA~\cite{zhu2022rethinking}, DI~\cite{CihangXie2018ImprovingTO}, we can see the whole image have been classified to one object in the segmentation map, indicating a successful attack. Similarly, for SegPGD~\cite{gu2022segpgd}, we can see there is no object in the segmentation map, which can be seen as a very successful attack.
Both Fig.~\ref{fig:model_performance_wrt_conventional_ad_attack} and Fig.~\ref{pascal_voc_attack_performance} indicate that deep semantic segmentation models are not robust to adversarial attack.

\noindent\textbf{Transferability of adversarial attack:} For a given network structure $F_i\in\{F_i\}_{i=1}^K$, where $K$ represents the number of model structures, we define it as the source model, and generate adversarial examples based on it to test robustness of other structures, e.g. $F_{j\neq i}$, with respect to the attacks based on $F_i$, thus we aim to evaluate transferability of adversarial attacks.

The success rate is analyzed to evaluate the transferability of adversarial attacks. In this paper, we have $K=3$ for each dataset-related model, and we report the success rate of each model on the Cityscape dataset in Fig.~\ref{fig:transferability_evaluation},
where we show the range of success rate of each attack. As a transferable attack, we aim to obtain an attack that can lead to a high success rate across different target models. Fig.~\ref{fig:transferability_evaluation} shows that FGSM~\cite{Explaining_and_Harnessing_Adversarial_Examples} and PGD~\cite{PGD_attack} attacks show almost consistent success rates across different target models, although the success rates are around 50\%. As a transferable model, we found DI~\cite{CihangXie2018ImprovingTO} fails to transfer well on the Cityscape dataset in the cross-network structure setting,
as its success rate is the lowest in general, however, its transferability on PascalVOC is reasonable, indicating the transferability should be discussed together with the dataset. Further, although DAG~\cite{xie2017adversarial} performs really well in the source model, its transferability on the Cityscape dataset is inferior compared with other conventional attacks. Among those attacks, NI~\cite{JiadongLin2019NesterovAG} still outperforms others to be the best transferable attack.
\cite{ilyas2019adversarial} has already shown that the image structure and feature bias will influence the transferability of the attack in the classification, which can be seen in our observation. Fig.~\ref{fig:transferability_evaluation} indicates that the transferability of adversarial attacks is a critical issue to evaluate model robustness. The current image classification based transferable attacks show limitations when they are directly adapted for semantic segmentation.

\begin{figure}[tp]
   \begin{center}
   \begin{tabular}{{c@{ }}}

{\includegraphics[width=0.95\linewidth]{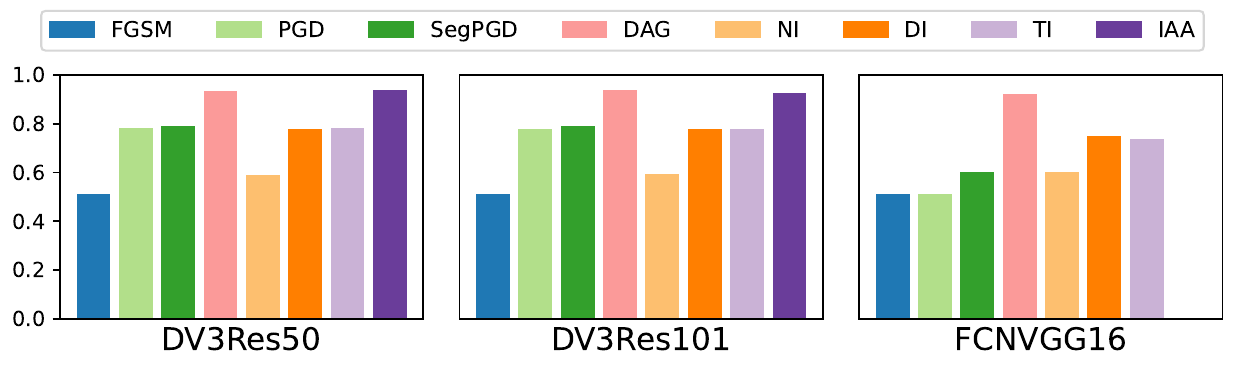}}\\
   \end{tabular}
   \end{center}
    \caption{SSIM of adversarial examples.
    (IAA~\cite{zhu2022rethinking} is designed for residual structure, thus it's not used in FCNVGG16.).
    } 
    \label{fig:model_ssim_wrt_conventional_ad_attack}
\end{figure}



\begin{figure}[t!]
   \begin{center}
   \begin{tabular}{{c@{ } }}
{\includegraphics[width=0.95\linewidth]{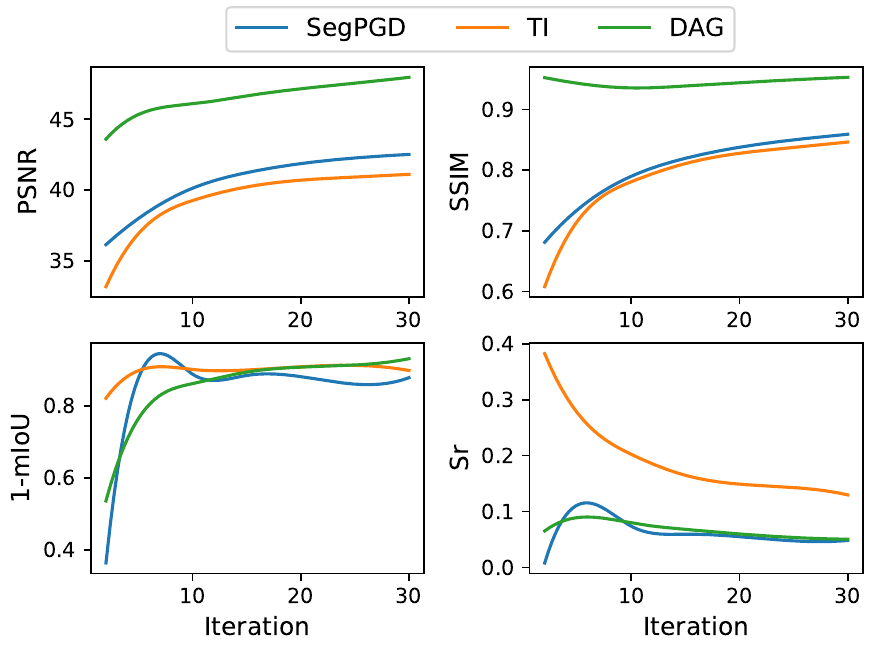}}
   \end{tabular}
   \end{center}
    \caption{Image quality (PSNR and SSIM) and model performance (1-mIOU and $S_r$) of adversarial examples \wrt~iterations.} 
    \label{image_quality_wrt_iteration}
\end{figure}

\subsection{Results Analysis}
\noindent\textbf{Image quality:}
We rank the adversarial attacks and transferable attacks based on their image quality (with SSIM) of the adversarial examples in Fig.~\ref{fig:model_ssim_wrt_conventional_ad_attack} on the Pascal VOC dataset. We argue that an effective adversarial attack should be visually invisible but significantly destroy model prediction, making adversarial image quality an important standard to decide the effectiveness of adversarial attack. Fig.~\ref{fig:model_ssim_wrt_conventional_ad_attack} shows that DAG~\cite{xie2017adversarial} leads to adversarial example with the highest $\text{SSIM}$, indicating its superiority in producing visually invisible attack. We argue one of the main reasons is that we set the iteration number to be small for fair comparison, where the adversarial example is still in an early stage of converging. Further, the final perturbation $\delta_{adv}$ of DAG~\cite{xie2017adversarial} is accumulated, which is then clipped before generating the adversarial example in practice, and both the above operations are beneficial for stable generating of the adversarial example. 


\noindent\textbf{Performance of iterative attacks \wrt~iteration:}
In this paper, we set the number of iterations to 10 to achieve a trade-off between attack quality and efficiency. We further evaluate attack performance \wrt~iteration,
and show the results in Fig.~\ref{image_quality_wrt_iteration}.
We observe that a larger iteration brings minor influence to the image quality of DAG~\cite{xie2017adversarial}, while it decreases image quality significantly for SegPGD~\cite{gu2022segpgd}. 
For the mIoU, since the attack is considered to be effective when mIoU is lower, we replace it with $1-\text{mIoU}$ to indicate the effectiveness of the attacks. It shows that the attack performance of SegPGD~\cite{gu2022segpgd} and DAG~\cite{xie2017adversarial} increase with the iteration going up at the beginning, but SegPGD~\cite{gu2022segpgd} goes stable after 10 iterations while DAG~\cite{xie2017adversarial} still slightly increases. The main reason for the saturated performance of SegPGD~\cite{gu2022segpgd} is that the misclassified samples are reducing, leading to nearly zero gradients. Another observation is that we find the stable performance of TI~\cite{YinpengDong2019EvadingDT} with increased
iterations, and we attribute this to the selection of the kernel matrix $\mathbf{W}$ in Eq.~\ref{eq_ti_pgd_attack}, where the selection of the pre-defined kernel matrix $\mathbf{W}$ is the bottleneck of transferability of TI~\cite{YinpengDong2019EvadingDT}.
   
\subsection{Ensemble attack}
We show performance of the proposed ensemble attack in Table~\ref{tab:pascalvoc_adversarial_attack_esm} and Table~\ref{tab:cityscape_adversarial_attack_esm} for Pascal VOC and Cityscape dataset, respectively. 
The experimental results show that the ensemble method performs better than the strongest single method (NI~\cite{JiadongLin2019NesterovAG}), leading to better transferability.
The basic idea of the ensemble of those three attacks is to use more augmentation of the data with more invariant features, and also utilize the accelerated gradient to avoid the attack coming into a local minimal and overfit. And more deeply, we are trying to find a generalized direction for the gradient descent.
Also, there is another way to the ensemble attack, where it ensemble multiple models and try to increase the transferability by using the multiple loss from the models.
However, since different models still can not cover all structure all the time, we adopt the first way of ensemble.




\begin{table*}[t!]
  \centering
  \footnotesize
  \renewcommand{\arraystretch}{1.2}
  \renewcommand{\tabcolsep}{4.25mm}
  \caption{Ensemble attack on Pascal VOC 2012 dataset.
  }
  \begin{tabular}{l|c|cc|cc|cc|cc}
  \toprule[1.1pt]
  &&\multicolumn{2}{c|}{Image Quality}&\multicolumn{6}{c}{Target (performance is evaluated with $\text{mIOU}\downarrow$ and $\text{Sr}\uparrow$)}\\ \hline
  Source & Attack & $\text{PSNR}\uparrow$ & $\text{SSIM}\uparrow$ & \multicolumn{2}{c|}{DV3Res50} & \multicolumn{2}{c|}{DV3Res101}& \multicolumn{2}{c}{FCNVGG16}  \\ \hline
  \multirow{2}{*}{DV3Res50} 
      &  NI~\cite{JiadongLin2019NesterovAG} & 32.84 & 0.5912 & 0.0861 & 0.8855 & 0.2610  &0.6650 & 0.4246 & 0.3478  \\ 
  & NI + DI + TI & 32.84 & \textbf{0.5961} & \textbf{0.0833} & \textbf{0.8893} & \textbf{0.2005} & \textbf{0.7427} & \textbf{0.3100} & \textbf{0.4657} \\
  
   \midrule[0.7pt]
   \multirow{2}{*}{DV3Res101}
    &  NI~\cite{JiadongLin2019NesterovAG} & 32.83 & 0.5925 & 0.2244 & 0.7017 & 0.0922 & 0.8817 & 0.4341  & 0.3332  \\ 
& NI + DI + TI & \textbf{32.85} & \textbf{0.5982} & \textbf{0.1920} & \textbf{0.7448} & \textbf{0.0855} & \textbf{0.8903} & \textbf{0.3256} & \textbf{0.4882} \\
  
\midrule[0.7pt]
  \multirow{2}{*}{FCNVGG16} 
   &  NI~\cite{JiadongLin2019NesterovAG} & \textbf{32.75} & 0.6018 & \textbf{0.1607} & \textbf{0.7864} & \textbf{0.2233} & \textbf{0.7134} &  0.0326 & 0.9499  \\ 
  & NI + DI + TI & 32.74 & \textbf{0.6072} & 0.1615 & 0.7853 & 0.2252 & 0.7109 & \textbf{0.0322} & \textbf{0.9505} \\
\bottomrule[1.1pt]
  \end{tabular}
  \label{tab:pascalvoc_adversarial_attack_esm}
\end{table*}

\begin{table*}[t!]
  \centering
  \footnotesize
  \renewcommand{\arraystretch}{1.2}
  \renewcommand{\tabcolsep}{4.25mm}
  \caption{Ensemble attack on Cityscape dataset.
  }
  \begin{tabular}{l|c|cc|cc|cc|cc}
  \toprule[1.1pt]
  &&\multicolumn{2}{c|}{Image Quality}&\multicolumn{6}{c}{Target (performance is evaluated with $\text{mIOU}\downarrow$ and $\text{Sr}\uparrow$)}\\ \hline
  Source & Attack & $\text{PSNR}\uparrow$ & $\text{SSIM}\uparrow$ & \multicolumn{2}{c|}{DV3Res50} & \multicolumn{2}{c|}{DV3Res101}& \multicolumn{2}{c}{FCNVGG16}  \\ \hline
  \multirow{2}{*}{DV3Res50} 
    & NI~\cite{JiadongLin2019NesterovAG}  & 32.91 & 0.3729 & 0.0142 & 0.9817 & 0.1398  & 0.8169 & 0.1754 & 0.7590\\ 
    
  & NI + DI + TI & \textbf{32.95} & \textbf{0.3822} & \textbf{0.0141} & \textbf{0.9818} & \textbf{0.0892} & \textbf{0.8832} & \textbf{0.0790} & \textbf{0.8914} \\
  
   \midrule[0.7pt]
   \multirow{2}{*}{DV3Res101}
    & NI~\cite{JiadongLin2019NesterovAG}  & 32.91 & 0.3736 & \textbf{0.1102} & \textbf{0.8578} & \textbf{0.0176} & \textbf{0.9770} &  0.1665 & 0.7712  \\ 
  & NI + DI + TI & \textbf{32.96} & \textbf{0.3855} & 0.1151 & 0.8515 & 0.0229 & 0.9700 & \textbf{0.1049} & \textbf{0.8559} \\
  
\midrule[0.7pt]
  \multirow{2}{*}{DV3MOB} 
    & NI~\cite{JiadongLin2019NesterovAG}  & 32.81 & 0.3739 & 0.2976 & 0.6160 & 0.3458 & 0.5471 &  0.0109 & 0.9850 \\ 
  & NI + DI + TI & \textbf{32.87} & \textbf{0.3854} & \textbf{0.1285} & \textbf{0.8342} & \textbf{0.2001} & \textbf{0.7380} & \textbf{0.0051} & \textbf{0.9930}\\
\bottomrule[1.1pt]
  \end{tabular}
  \label{tab:cityscape_adversarial_attack_esm}
\end{table*}



\begin{figure}[tp]
   \begin{center}
   \begin{tabular}{{c@{ } c@{ } c@{ }}}
{\includegraphics[width=0.3\linewidth]{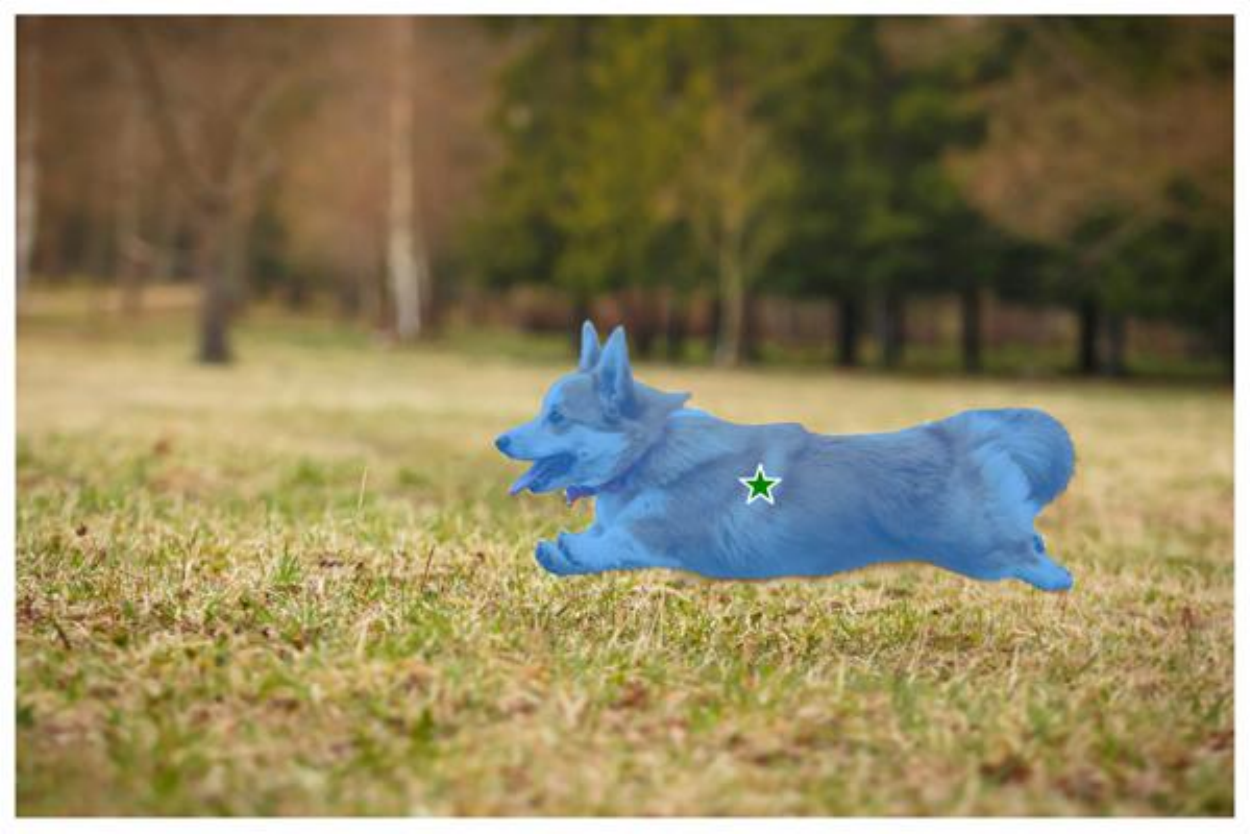}}&  
{\includegraphics[width=0.3\linewidth]{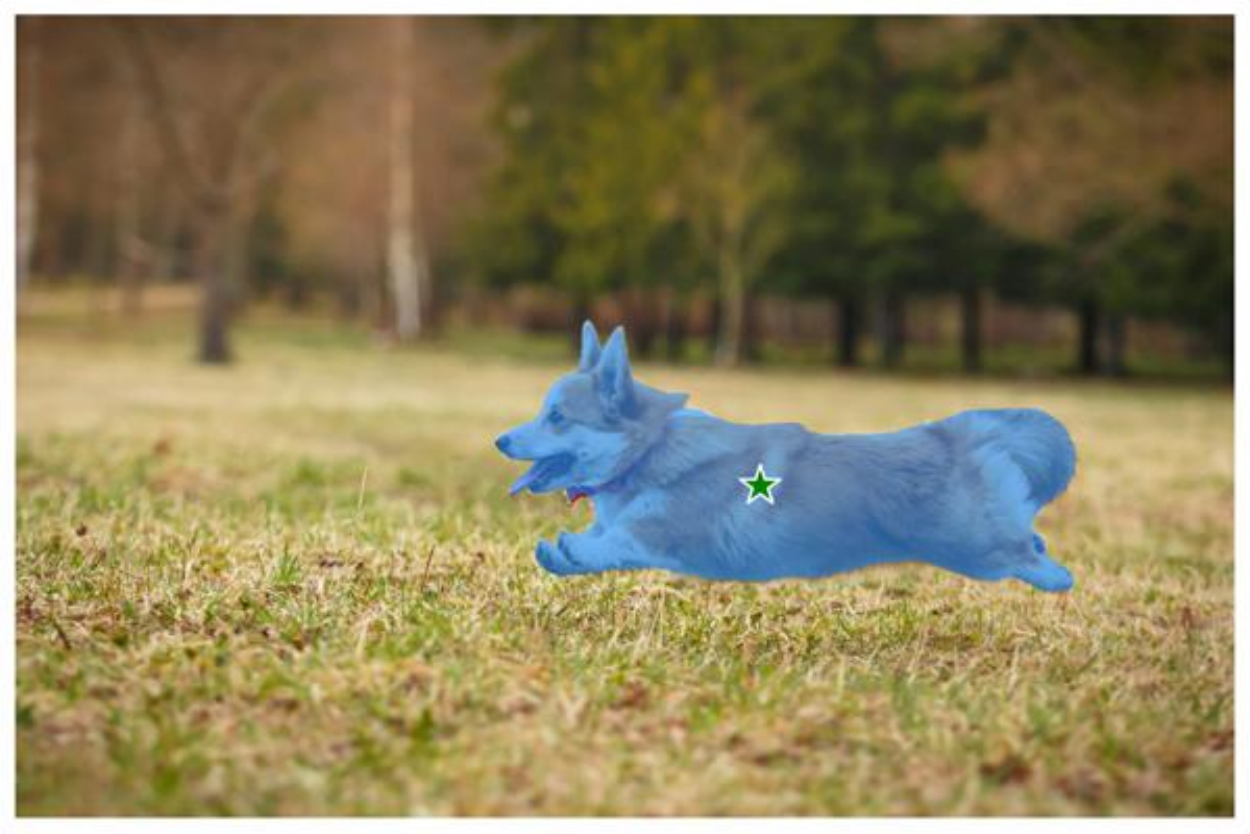}}&  
{\includegraphics[width=0.3\linewidth]{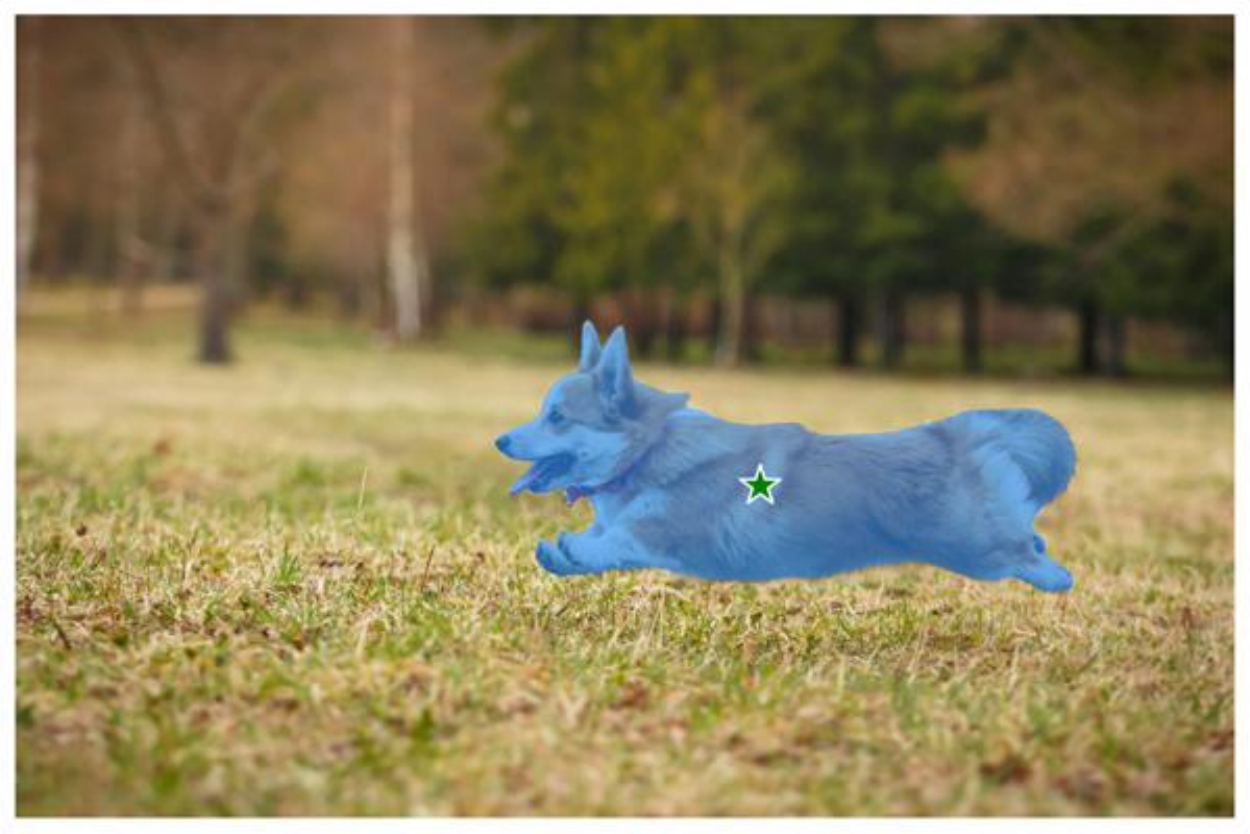}}\\ 
\footnotesize{ViT-B Clean}&\footnotesize{ViT-L Clean}&\footnotesize{ViT-H Clean}\\
{\includegraphics[width=0.3\linewidth]{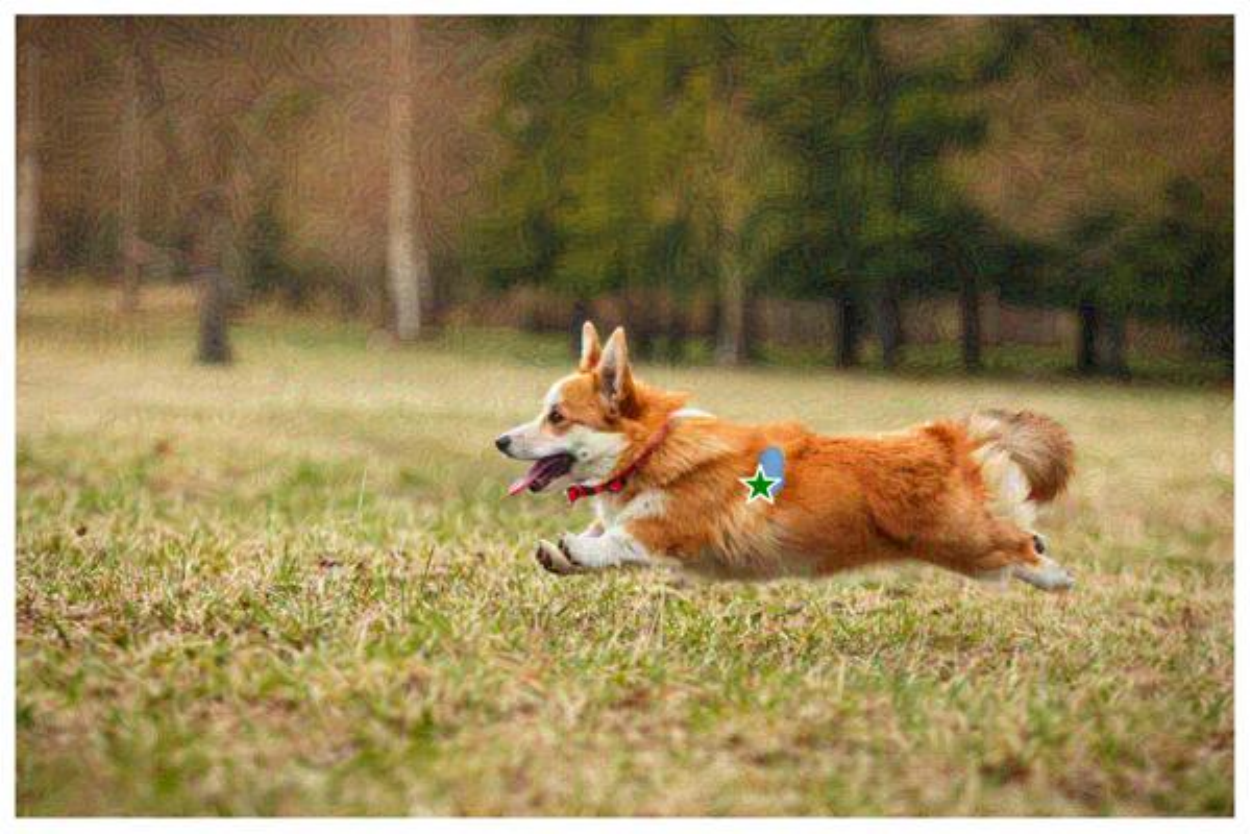}}&  
{\includegraphics[width=0.3\linewidth]{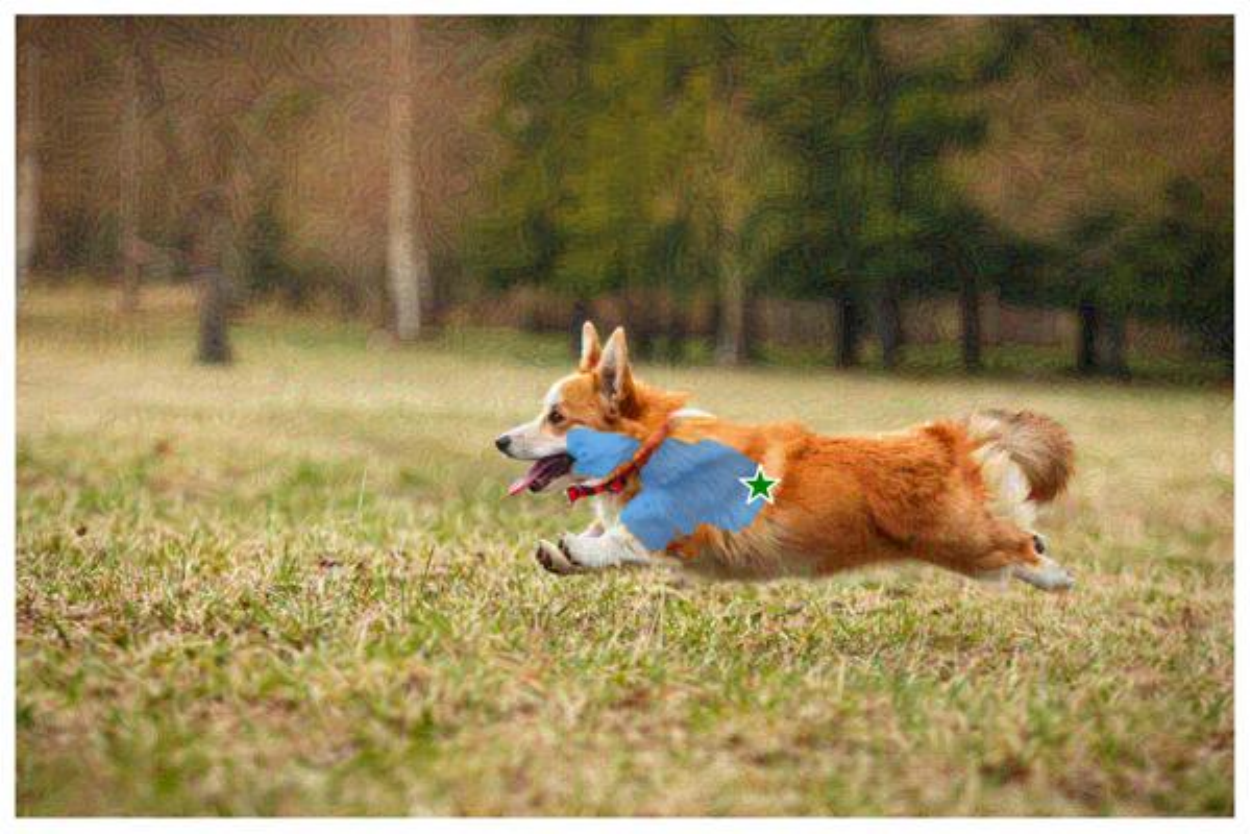}}&  
{\includegraphics[width=0.3\linewidth]{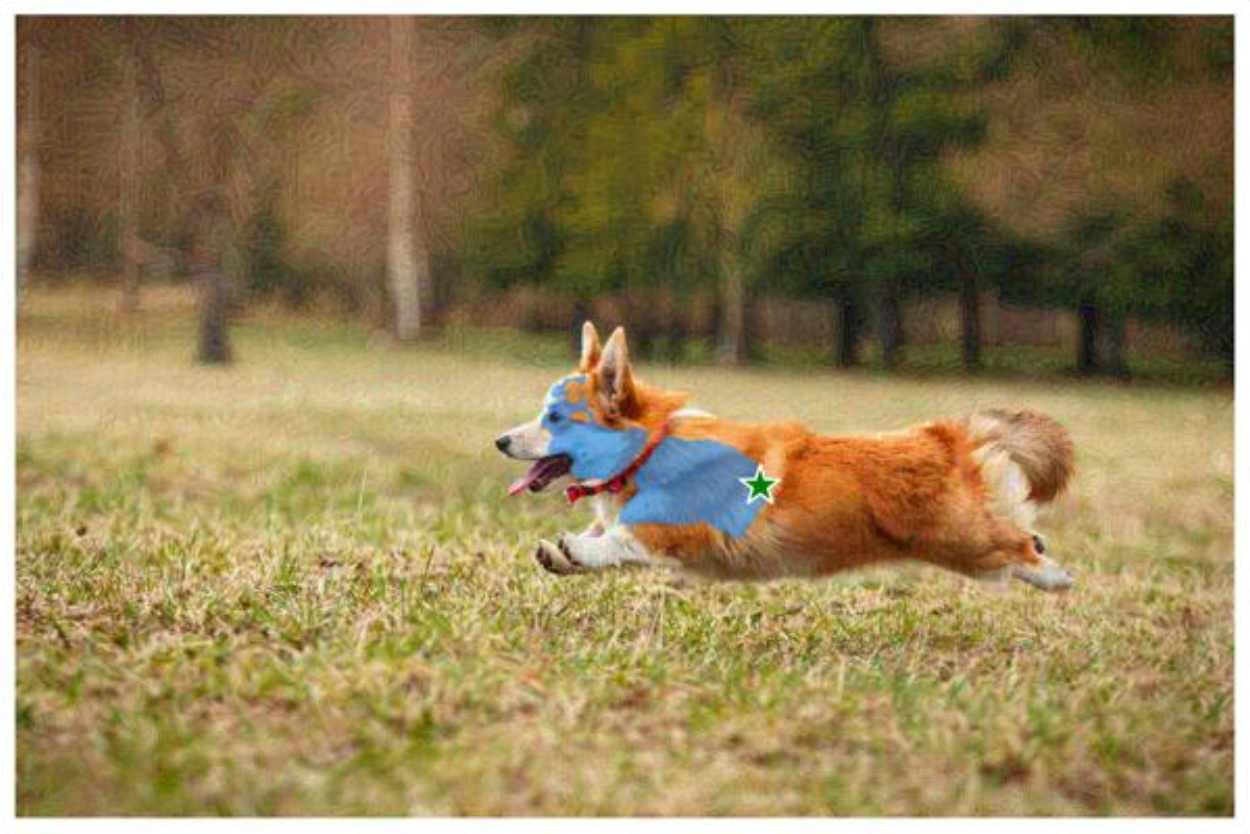}}\\ 
\footnotesize{ViT-B\_NI}&\footnotesize{ViT-L\_NI}&\footnotesize{ViT-H\_NI}\\
{\includegraphics[width=0.3\linewidth]{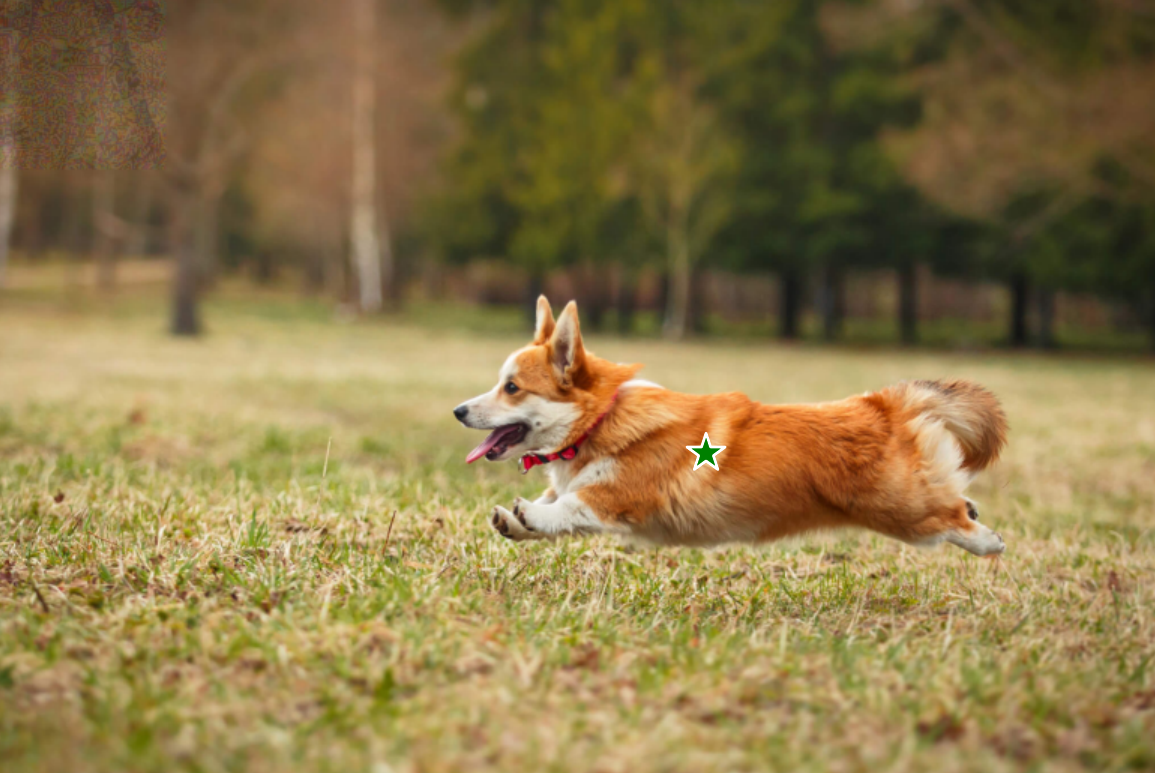}}&  
{\includegraphics[width=0.3\linewidth]{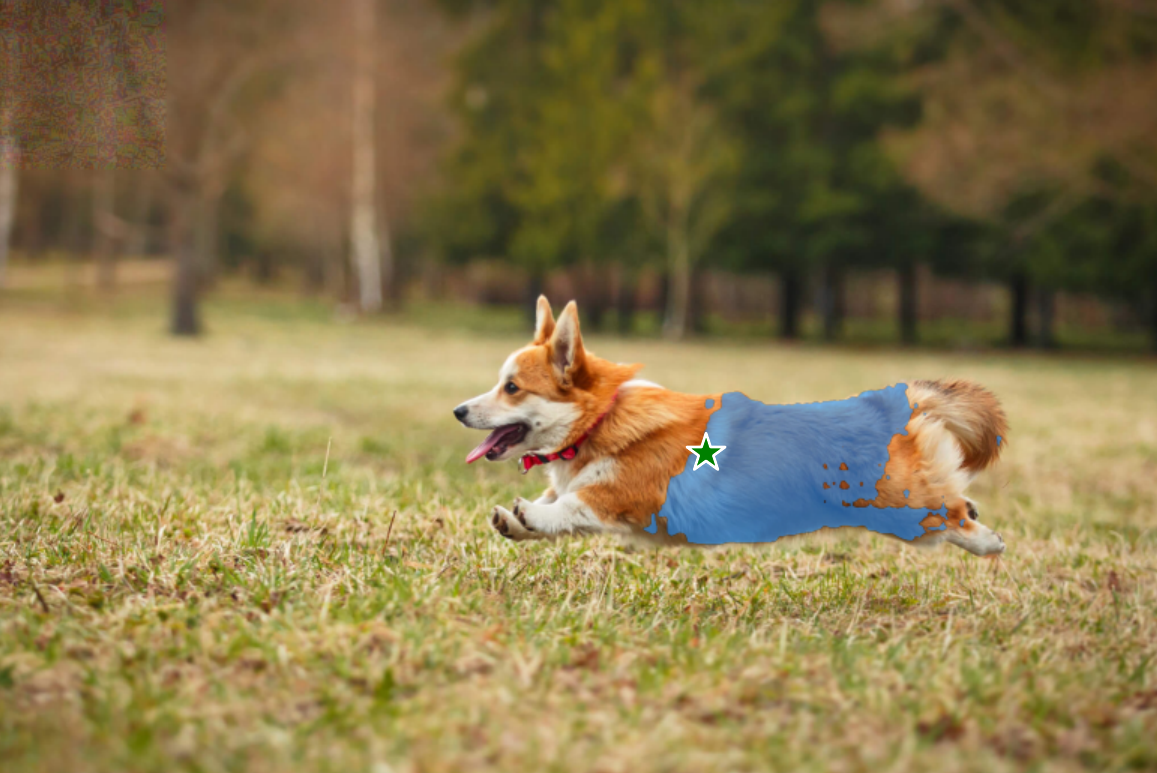}}&  
{\includegraphics[width=0.3\linewidth]{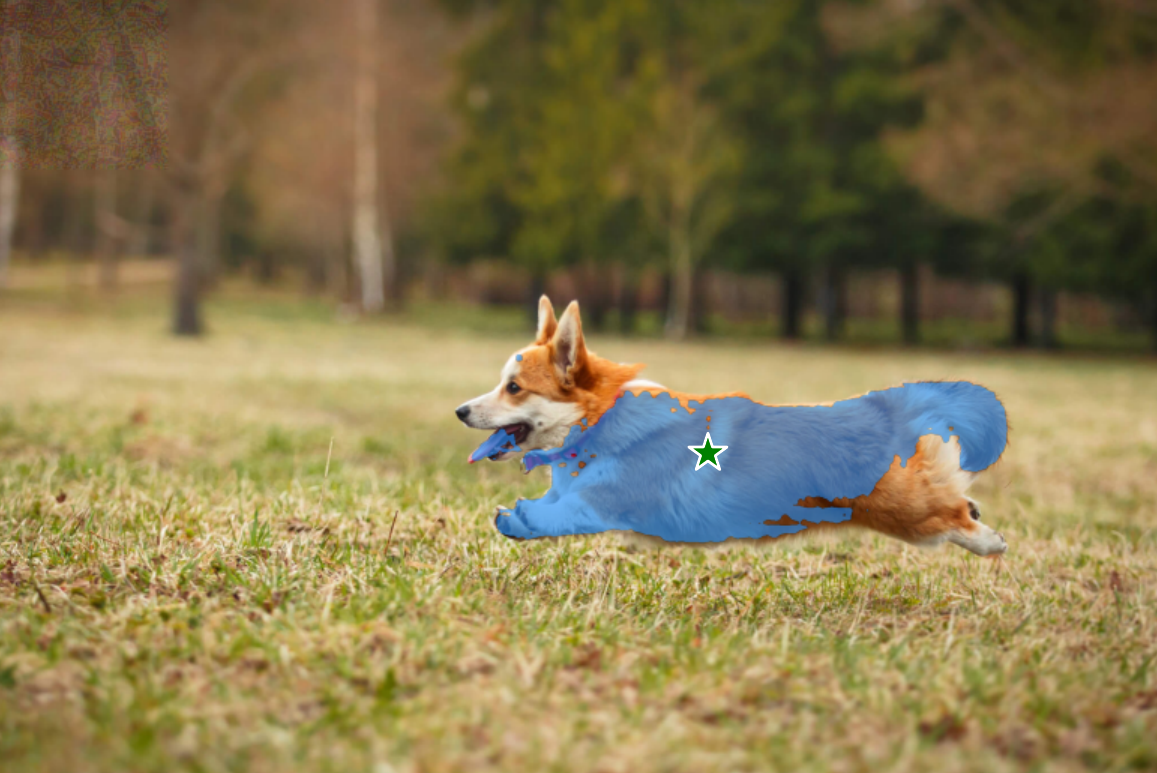}}\\ 
\footnotesize{ViT-B\_Ensemble}&\footnotesize{ViT-L\_Ensemble}&\footnotesize{ViT-H\_Ensemble}\\
   \end{tabular}
   \end{center}
    \caption{Transferable adversarial attack for Segment Anything~\cite{kirillov2023segany} with point prompts, where the top row shows predictions of three backbones on the clean RGB image. 
    The second and third rows show predictions of three backbones on the adversarial image generated from ViT-B (the source model) by NI~\cite{JiadongLin2019NesterovAG} and the proposed ensemble attack, respectively.
    } 
    \label{transfer_attack_sam}
\end{figure}

\noindent\textbf{Transferable attack of existing vision foundation models:}
With recent advancement of vision foundation models such as SAM~\cite{kirillov2023segany}, CLIP~\cite{DBLP:conf/icml/RadfordKHRGASAM21}, ImageBind~\cite{girdhar2023imagebind} and~\etc, more attentions have been paid to the transformer-based segmentation models. Compared with convolutional networks, generating adversarial examples through large transformer models is computationally expensive. Thus, if adversarial examples generated from a smaller model can transfer to larger models, we could significantly reduce the cost of generating adversarial examples for large vision foundation models. 

We perform a preliminary experiment on Segment Anything (SAM)~\cite{kirillov2023segany} with point prompts and show the result in Fig.~\ref{transfer_attack_sam}. To be specific, we consider three network architectures from SAM: ViT-B (86M), ViT-L (307M) as well as ViT-H (632M)~\cite{li2022exploring}, where the number after each backbone indicates the number of parameters. We use the smallest backbone ViT-B as the source model to generate the adversarial example using the proposed ensemble attack as well as NI
with a noise budget of $12/255$ and 16 iterations.
The \enquote{Clean} segmentation results represent predictions of the visual foundation models with clean RGB image as input. \enquote{Source} indicates the source model to generate adversarial attack, namely
ViT-B in our case.
We then evaluate the robustness of the other two \enquote{Target} models by feeding the adversarial example from the source model to the corresponding target model.

The missing detection of the foreground object (\enquote{ViT-B (Source)}) explains that existing vision foundation models are also vulnerable to adversarial attacks. 
Further, we observe that the adversarial example generated from the ViT-B is able to generalize to ViT-L as well as ViT-H and the attack remains effective for different point prompts. 
Thus, with the proposed ensemble attack as well as NI~\cite{JiadongLin2019NesterovAG}, we can generate adversarial examples from smaller models and transfer them to larger models with better computational efficiency. 
We also discover that the proposed ensemble attack has sub-optimal transferability than NI~\cite{JiadongLin2019NesterovAG}. We conjecture that current ensemble-based transfer attack methods are designed for the CNN, thus, they have better performance on CNN based segmentation model than the transformer-based segmentation model. More investigation on transferable attack for transformer backbone based models will be conducted.
\section{Conclusion}
We investigate transferability attack for semantic segmentation,
\ie~adversarial attack across different network structures. We observe that semantic segmentation models are vulnerable to adversarial attack.
We find that directly enforcing the model to produce pixel-wise wrong prediction can produce effective attack for the source model, \ie~DAG~~\cite{xie2017adversarial},
but it cannot transfer well to other models of different structures, indicating lower transferability. As a transferable attack, NI~\cite{JiadongLin2019NesterovAG} achieves the best attack transferability, however, its sample quality is inferior to DAG~\cite{xie2017adversarial}. Table~\ref{tab:pascalvoc_adversarial_attack} and \ref{tab:adversarial_cityscape} also reveal that adversarial attack generated from larger model (see Table~\ref{tab:backbone_parameter_numbers}) can transfer better than that from smaller models, \ie~DV3MOB~\cite{sandler2018mobilenetv2}.
We also identify the easier path to generate transferable attack is defining the source model as a residual connected based one, which transfers better to non-residual counterparts than the other way around.
We find dataset is also an important issue, where 
IAA~\cite{zhu2022rethinking} explores transferable attack from data distribution perspective. However, due to the implementation gap, our results fail to validate its effectiveness. 

We believe directly targeting wrong predictions, considering both network structure differences and data distribution gaps should be the top three factors for transferable attack. Further, our preliminary investigation on ensemble attack and transferable attack for vision foundation model reveals the potential of combining different attacks for a more robust attack, and the vulnerability of existing vision foundation models, strengthening the necessity for research on transferable attacks. More research will be conducted to theoretically understand attack transferability \wrt~dataset distribution, model structure and etc., especially robustness of vision foundation models \wrt~invisible input perturbation to achieve safe model deployment.
{
\bibliographystyle{IEEEtran}
\bibliography{egbib}

\begin{thebibliography}{10}
\providecommand{\url}[1]{#1}
\csname url@samestyle\endcsname
\providecommand{\newblock}{\relax}
\providecommand{\bibinfo}[2]{#2}
\providecommand{\BIBentrySTDinterwordspacing}{\spaceskip=0pt\relax}
\providecommand{\BIBentryALTinterwordstretchfactor}{4}
\providecommand{\BIBentryALTinterwordspacing}{\spaceskip=\fontdimen2\font plus
\BIBentryALTinterwordstretchfactor\fontdimen3\font minus
  \fontdimen4\font\relax}
\providecommand{\BIBforeignlanguage}[2]{{%
\expandafter\ifx\csname l@#1\endcsname\relax
\typeout{** WARNING: IEEEtran.bst: No hyphenation pattern has been}%
\typeout{** loaded for the language `#1'. Using the pattern for}%
\typeout{** the default language instead.}%
\else
\language=\csname l@#1\endcsname
\fi
#2}}
\providecommand{\BIBdecl}{\relax}
\BIBdecl

\bibitem{Cordts2016Cityscapes}
M.~Cordts, M.~Omran, S.~Ramos, T.~Rehfeld, M.~Enzweiler, R.~Benenson,
  U.~Franke, S.~Roth, and B.~Schiele, ``The cityscapes dataset for semantic
  urban scene understanding,'' in \emph{Proc. of the IEEE Conference on
  Computer Vision and Pattern Recognition (CVPR)}, 2016.

\bibitem{chen2017rethinking}
\BIBentryALTinterwordspacing
L.~Chen, G.~Papandreou, F.~Schroff, and H.~Adam, ``Rethinking atrous
  convolution for semantic image segmentation,'' \emph{CoRR}, vol.
  abs/1706.05587, 2017. [Online]. Available:
  \url{http://arxiv.org/abs/1706.05587}
\BIBentrySTDinterwordspacing

\bibitem{PGD_attack}
\BIBentryALTinterwordspacing
A.~Madry, A.~Makelov, L.~Schmidt, D.~Tsipras, and A.~Vladu, ``Towards deep
  learning models resistant to adversarial attacks,'' 2017. [Online].
  Available: \url{https://arxiv.org/abs/1706.06083}
\BIBentrySTDinterwordspacing

\bibitem{chen2018encoder}
L.-C. Chen, Y.~Zhu, G.~Papandreou, F.~Schroff, and H.~Adam, ``Encoder-decoder
  with atrous separable convolution for semantic image segmentation,'' in
  \emph{Proceedings of the European conference on computer vision (ECCV)},
  2018, pp. 801--818.

\bibitem{Explaining_and_Harnessing_Adversarial_Examples}
I.~J. Goodfellow, J.~Shlens, and C.~Szegedy, ``Explaining and harnessing
  adversarial examples,'' in \emph{International Conference on Learning
  Representations (ICLR)}, 2015.

\bibitem{chen2022vision}
R.~Ranftl, A.~Bochkovskiy, and V.~Koltun, ``Vision transformers for dense
  prediction,'' in \emph{Proceedings of the IEEE/CVF International Conference
  on Computer Vision (ICCV)}, October 2021, pp. 12\,179--12\,188.

\bibitem{yuan2019segmentation}
\BIBentryALTinterwordspacing
Y.~Yuan, X.~Chen, and J.~Wang, ``Object-contextual representations for semantic
  segmentation,'' in \emph{Computer Vision - {ECCV} 2020 - 16th European
  Conference, Glasgow, UK, August 23-28, 2020, Proceedings, Part {VI}}, ser.
  Lecture Notes in Computer Science, A.~Vedaldi, H.~Bischof, T.~Brox, and
  J.~Frahm, Eds., vol. 12351.\hskip 1em plus 0.5em minus 0.4em\relax Springer,
  2020, pp. 173--190. [Online]. Available:
  \url{https://doi.org/10.1007/978-3-030-58539-6\_11}
\BIBentrySTDinterwordspacing

\bibitem{yan2022lawin}
\BIBentryALTinterwordspacing
H.~Yan, C.~Zhang, and M.~Wu, ``Lawin transformer: Improving semantic
  segmentation transformer with multi-scale representations via large window
  attention,'' \emph{CoRR}, vol. abs/2201.01615, 2022. [Online]. Available:
  \url{https://arxiv.org/abs/2201.01615}
\BIBentrySTDinterwordspacing

\bibitem{mohan2021efficientps}
R.~Mohan and A.~Valada, ``Efficientps: Efficient panoptic segmentation,''
  \emph{International Journal of Computer Vision}, vol. 129, no.~5, pp.
  1551--1579, 2021.

\bibitem{cheng2020panoptic}
B.~Cheng, M.~D. Collins, Y.~Zhu, T.~Liu, T.~S. Huang, H.~Adam, and L.-C. Chen,
  ``Panoptic-deeplab: A simple, strong, and fast baseline for bottom-up
  panoptic segmentation,'' in \emph{Proceedings of the IEEE/CVF conference on
  computer vision and pattern recognition}, 2020, pp. 12\,475--12\,485.

\bibitem{zhang2021dcnas}
X.~Zhang, H.~Xu, H.~Mo, J.~Tan, C.~Yang, L.~Wang, and W.~Ren, ``Dcnas: Densely
  connected neural architecture search for semantic image segmentation,'' in
  \emph{Proceedings of the IEEE/CVF conference on computer vision and pattern
  recognition}, 2021, pp. 13\,956--13\,967.

\bibitem{li2019global}
X.~Li, L.~Zhang, G.~Cheng, K.~Yang, Y.~Tong, X.~Zhu, and T.~Xiang, ``Global
  aggregation then local distribution for scene parsing,'' \emph{IEEE
  Transactions on Image Processing}, vol.~30, pp. 6829--6842, 2021.

\bibitem{long2015fully}
J.~Long, E.~Shelhamer, and T.~Darrell, ``Fully convolutional networks for
  semantic segmentation,'' in \emph{Proceedings of the IEEE conference on
  computer vision and pattern recognition}, 2015, pp. 3431--3440.

\bibitem{lin2017refinenet}
G.~Lin, A.~Milan, C.~Shen, and I.~Reid, ``Refinenet: Multi-path refinement
  networks for high-resolution semantic segmentation,'' in \emph{Proceedings of
  the IEEE conference on computer vision and pattern recognition}, 2017, pp.
  1925--1934.

\bibitem{chen2016attention}
L.-C. Chen, Y.~Yang, J.~Wang, W.~Xu, and A.~L. Yuille, ``Attention to scale:
  Scale-aware semantic image segmentation,'' in \emph{Proceedings of the IEEE
  conference on computer vision and pattern recognition}, 2016, pp. 3640--3649.

\bibitem{acuna2019devil}
D.~Acuna, A.~Kar, and S.~Fidler, ``Devil is in the edges: Learning semantic
  boundaries from noisy annotations,'' in \emph{Proceedings of the IEEE/CVF
  conference on computer vision and pattern recognition}, 2019, pp.
  11\,075--11\,083.

\bibitem{girshick2014rich}
R.~Girshick, J.~Donahue, T.~Darrell, and J.~Malik, ``Rich feature hierarchies
  for accurate object detection and semantic segmentation,'' in
  \emph{Proceedings of the IEEE conference on computer vision and pattern
  recognition}, 2014, pp. 580--587.

\bibitem{chen2017deeplab}
L.-C. Chen, G.~Papandreou, I.~Kokkinos, K.~Murphy, and A.~L. Yuille, ``Deeplab:
  Semantic image segmentation with deep convolutional nets, atrous convolution,
  and fully connected crfs,'' 2017.

\bibitem{deng2019restricted}
L.~Deng, M.~Yang, H.~Li, T.~Li, B.~Hu, and C.~Wang, ``Restricted deformable
  convolution-based road scene semantic segmentation using surround view
  cameras,'' \emph{IEEE Transactions on Intelligent Transportation Systems},
  vol.~21, no.~10, pp. 4350--4362, 2019.

\bibitem{goodfellow2020generative}
I.~Goodfellow, J.~Pouget-Abadie, M.~Mirza, B.~Xu, D.~Warde-Farley, S.~Ozair,
  A.~Courville, and Y.~Bengio, ``Generative adversarial networks,''
  \emph{Communications of the ACM}, vol.~63, no.~11, pp. 139--144, 2020.

\bibitem{kingma2013auto}
D.~P. Kingma and M.~Welling, ``{Auto-Encoding Variational Bayes},'' in
  \emph{2nd International Conference on Learning Representations, {ICLR} 2014,
  Banff, AB, Canada, April 14-16, 2014, Conference Track Proceedings}, 2014.

\bibitem{li2021semantic}
D.~Li, J.~Yang, K.~Kreis, A.~Torralba, and S.~Fidler, ``Semantic segmentation
  with generative models: Semi-supervised learning and strong out-of-domain
  generalization,'' in \emph{2021 IEEE/CVF Conference on Computer Vision and
  Pattern Recognition (CVPR)}, 2021, pp. 8296--8307.

\bibitem{souly2017semi}
N.~Souly, C.~Spampinato, and M.~Shah, ``Semi supervised semantic segmentation
  using generative adversarial network,'' in \emph{2017 IEEE International
  Conference on Computer Vision (ICCV)}, 2017, pp. 5689--5697.

\bibitem{luc2016semantic}
\BIBentryALTinterwordspacing
P.~Luc, C.~Couprie, S.~Chintala, and J.~Verbeek, ``Semantic segmentation using
  adversarial networks,'' \emph{CoRR}, vol. abs/1611.08408, 2016. [Online].
  Available: \url{http://arxiv.org/abs/1611.08408}
\BIBentrySTDinterwordspacing

\bibitem{xue2018segan}
Y.~Xue, T.~Xu, H.~Zhang, L.~R. Long, and X.~Huang, ``Segan: Adversarial network
  with multi-scale l 1 loss for medical image segmentation,''
  \emph{Neuroinformatics}, vol.~16, pp. 383--392, 2018.

\bibitem{zhaoa2021semantic}
Z.~Zhaoa, Y.~Wang, K.~Liu, H.~Yang, Q.~Sun, and H.~Qiao, ``Semantic
  segmentation by improved generative adversarial networks,'' 2021.

\bibitem{xie2017adversarial}
C.~Xie, J.~Wang, Z.~Zhang, Y.~Zhou, L.~Xie, and A.~Yuille, ``Adversarial
  examples for semantic segmentation and object detection,'' in \emph{2017 IEEE
  International Conference on Computer Vision (ICCV)}, 2017, pp. 1378--1387.

\bibitem{arnab2018robustness}
A.~Arnab, O.~Miksik, and P.~H. Torr, ``On the robustness of semantic
  segmentation models to adversarial attacks,'' in \emph{Proceedings of the
  IEEE Conference on Computer Vision and Pattern Recognition}, 2018, pp.
  888--897.

\bibitem{gu2022segpgd}
J.~Gu, H.~Zhao, V.~Tresp, and P.~H. Torr, ``Segpgd: An effective and efficient
  adversarial attack for evaluating and boosting segmentation robustness,'' in
  \emph{Computer Vision--ECCV 2022: 17th European Conference, Tel Aviv, Israel,
  October 23--27, 2022, Proceedings, Part XXIX}.\hskip 1em plus 0.5em minus
  0.4em\relax Springer, 2022, pp. 308--325.

\bibitem{kirillov2023segany}
\BIBentryALTinterwordspacing
A.~Kirillov, E.~Mintun, N.~Ravi, H.~Mao, C.~Rolland, L.~Gustafson, T.~Xiao,
  S.~Whitehead, A.~C. Berg, W.~Lo, P.~Doll{\'{a}}r, and R.~B. Girshick,
  ``Segment anything,'' \emph{CoRR}, vol. abs/2304.02643, 2023. [Online].
  Available: \url{https://doi.org/10.48550/arXiv.2304.02643}
\BIBentrySTDinterwordspacing

\bibitem{szegedy2013intriguing}
\BIBentryALTinterwordspacing
C.~Szegedy, W.~Zaremba, I.~Sutskever, J.~Bruna, D.~Erhan, I.~J. Goodfellow, and
  R.~Fergus, ``Intriguing properties of neural networks,'' in \emph{2nd
  International Conference on Learning Representations, {ICLR} 2014, Banff, AB,
  Canada, April 14-16, 2014, Conference Track Proceedings}, Y.~Bengio and
  Y.~LeCun, Eds., 2014. [Online]. Available:
  \url{http://arxiv.org/abs/1312.6199}
\BIBentrySTDinterwordspacing

\bibitem{Towards_Evaluating_Robustness_Neural_Networks}
\BIBentryALTinterwordspacing
N.~Carlini and D.~A. Wagner, ``Towards evaluating the robustness of neural
  networks,'' \emph{CoRR}, vol. abs/1608.04644, 2016. [Online]. Available:
  \url{http://arxiv.org/abs/1608.04644}
\BIBentrySTDinterwordspacing

\bibitem{Ensemble_Adversarial_Training_Attacks_Defenses}
\BIBentryALTinterwordspacing
F.~Tramèr, A.~Kurakin, N.~Papernot, I.~Goodfellow, D.~Boneh, and P.~McDaniel,
  ``Ensemble adversarial training: Attacks and defenses,'' 2017. [Online].
  Available: \url{https://arxiv.org/abs/1705.07204}
\BIBentrySTDinterwordspacing

\bibitem{DeepFool}
\BIBentryALTinterwordspacing
S.~Moosavi{-}Dezfooli, A.~Fawzi, and P.~Frossard, ``Deepfool: {A} simple and
  accurate method to fool deep neural networks,'' in \emph{2016 {IEEE}
  Conference on Computer Vision and Pattern Recognition, {CVPR} 2016, Las
  Vegas, NV, USA, June 27-30, 2016}.\hskip 1em plus 0.5em minus 0.4em\relax
  {IEEE} Computer Society, 2016, pp. 2574--2582. [Online]. Available:
  \url{https://doi.org/10.1109/CVPR.2016.282}
\BIBentrySTDinterwordspacing

\bibitem{Takikawa_2019_ICCV}
T.~Takikawa, D.~Acuna, V.~Jampani, and S.~Fidler, ``Gated-scnn: Gated shape
  cnns for semantic segmentation,'' in \emph{Proceedings of the IEEE/CVF
  International Conference on Computer Vision (ICCV)}, October 2019.

\bibitem{yuan2020segfix}
Y.~Yuan, J.~Xie, X.~Chen, and J.~Wang, ``Segfix: Model-agnostic boundary
  refinement for segmentation,'' 2020.

\bibitem{Liang_2020_CVPR}
J.~Liang, N.~Homayounfar, W.-C. Ma, Y.~Xiong, R.~Hu, and R.~Urtasun,
  ``Polytransform: Deep polygon transformer for instance segmentation,'' in
  \emph{Proceedings of the IEEE/CVF Conference on Computer Vision and Pattern
  Recognition (CVPR)}, June 2020.

\bibitem{wang2018understanding}
P.~Wang, P.~Chen, Y.~Yuan, D.~Liu, Z.~Huang, X.~Hou, and G.~Cottrell,
  ``Understanding convolution for semantic segmentation,'' in \emph{2018 IEEE
  winter conference on applications of computer vision (WACV)}.\hskip 1em plus
  0.5em minus 0.4em\relax Ieee, 2018, pp. 1451--1460.

\bibitem{zhao2017pyramid}
H.~Zhao, J.~Shi, X.~Qi, X.~Wang, and J.~Jia, ``Pyramid scene parsing network,''
  in \emph{Proceedings of the IEEE conference on computer vision and pattern
  recognition}, 2017, pp. 2881--2890.

\bibitem{dai2017deformable}
J.~Dai, H.~Qi, Y.~Xiong, Y.~Li, G.~Zhang, H.~Hu, and Y.~Wei, ``Deformable
  convolutional networks,'' in \emph{Proceedings of the IEEE international
  conference on computer vision}, 2017, pp. 764--773.

\bibitem{liu2015semantic}
Z.~Liu, X.~Li, P.~Luo, C.-C. Loy, and X.~Tang, ``Semantic image segmentation
  via deep parsing network,'' in \emph{Proceedings of the IEEE international
  conference on computer vision}, 2015, pp. 1377--1385.

\bibitem{lin2016efficient}
G.~Lin, C.~Shen, A.~Van Den~Hengel, and I.~Reid, ``Efficient piecewise training
  of deep structured models for semantic segmentation,'' in \emph{Proceedings
  of the IEEE conference on computer vision and pattern recognition}, 2016, pp.
  3194--3203.

\bibitem{krahenbuhl2011efficient}
P.~Kr{\"a}henb{\"u}hl and V.~Koltun, ``Efficient inference in fully connected
  crfs with gaussian edge potentials,'' in \emph{Advances in neural information
  processing systems}, vol.~24, 2011.

\bibitem{vaswani2017attention}
A.~Vaswani, N.~Shazeer, N.~Parmar, J.~Uszkoreit, L.~Jones, A.~N. Gomez,
  {\L}.~Kaiser, and I.~Polosukhin, ``Attention is all you need,''
  \emph{Advances in neural information processing systems}, vol.~30, 2017.

\bibitem{chen2021transunet}
\BIBentryALTinterwordspacing
J.~Chen, Y.~Lu, Q.~Yu, X.~Luo, E.~Adeli, Y.~Wang, L.~Lu, A.~L. Yuille, and
  Y.~Zhou, ``Transunet: Transformers make strong encoders for medical image
  segmentation,'' \emph{CoRR}, vol. abs/2102.04306, 2021. [Online]. Available:
  \url{https://arxiv.org/abs/2102.04306}
\BIBentrySTDinterwordspacing

\bibitem{strudel2021segmenter}
R.~Strudel, R.~Garcia, I.~Laptev, and C.~Schmid, ``Segmenter: Transformer for
  semantic segmentation,'' 2021.

\bibitem{ranftl2021vision}
R.~Ranftl, A.~Bochkovskiy, and V.~Koltun, ``Vision transformers for dense
  prediction,'' in \emph{Proceedings of the IEEE/CVF international conference
  on computer vision}, 2021, pp. 12\,179--12\,188.

\bibitem{zheng2021rethinking}
S.~Zheng, J.~Lu, H.~Zhao, X.~Zhu, Z.~Luo, Y.~Wang, Y.~Fu, J.~Feng, T.~Xiang,
  P.~H.~S. Torr, and L.~Zhang, ``Rethinking semantic segmentation from a
  sequence-to-sequence perspective with transformers,'' 2021.

\bibitem{xie2021segformer}
E.~Xie, W.~Wang, Z.~Yu, A.~Anandkumar, J.~M. Alvarez, and P.~Luo, ``Segformer:
  Simple and efficient design for semantic segmentation with transformers,''
  2021.

\bibitem{cheng2021perpixel}
B.~Cheng, A.~Schwing, and A.~Kirillov, ``Per-pixel classification is not all
  you need for semantic segmentation,'' in \emph{Advances in Neural Information
  Processing Systems}, 2021, pp. 17\,864--17\,875.

\bibitem{cheng2022masked}
B.~Cheng, I.~Misra, A.~G. Schwing, A.~Kirillov, and R.~Girdhar,
  ``Masked-attention mask transformer for universal image segmentation,'' in
  \emph{Proceedings of the IEEE/CVF conference on computer vision and pattern
  recognition}, 2022, pp. 1290--1299.

\bibitem{hendrik2017universal}
J.~Hendrik~Metzen, M.~Chaithanya~Kumar, T.~Brox, and V.~Fischer, ``Universal
  adversarial perturbations against semantic image segmentation,'' in
  \emph{Proceedings of the IEEE international conference on computer vision},
  2017, pp. 2755--2764.

\bibitem{fischer2017adversarial}
\BIBentryALTinterwordspacing
V.~Fischer, M.~C. Kumar, J.~H. Metzen, and T.~Brox, ``Adversarial examples for
  semantic image segmentation,'' \emph{CoRR}, vol. abs/1703.01101, 2017.
  [Online]. Available: \url{http://arxiv.org/abs/1703.01101}
\BIBentrySTDinterwordspacing

\bibitem{moosavi2017universal}
S.-M. Moosavi-Dezfooli, A.~Fawzi, O.~Fawzi, and P.~Frossard, ``Universal
  adversarial perturbations,'' in \emph{Proceedings of the IEEE conference on
  computer vision and pattern recognition}, 2017, pp. 1765--1773.

\bibitem{adversarial_example_semantic_seg_iclr2017workshop}
\BIBentryALTinterwordspacing
V.~Fischer, M.~C. Kumar, J.~H. Metzen, and T.~Brox, ``Adversarial examples for
  semantic image segmentation,'' 2017. [Online]. Available:
  \url{https://arxiv.org/abs/1703.01101}
\BIBentrySTDinterwordspacing

\bibitem{li2021hidden}
\BIBentryALTinterwordspacing
Y.~Li, Y.~Li, Y.~Lv, Y.~Jiang, and S.~Xia, ``Hidden backdoor attack against
  semantic segmentation models,'' \emph{CoRR}, vol. abs/2103.04038, 2021.
  [Online]. Available: \url{https://arxiv.org/abs/2103.04038}
\BIBentrySTDinterwordspacing

\bibitem{agnihotri2023cospgd}
\BIBentryALTinterwordspacing
S.~Agnihotri and M.~Keuper, ``Cospgd: a unified white-box adversarial attack
  for pixel-wise prediction tasks,'' \emph{CoRR}, vol. abs/2302.02213, 2023.
  [Online]. Available: \url{https://doi.org/10.48550/arXiv.2302.02213}
\BIBentrySTDinterwordspacing

\bibitem{xu2021dynamic}
X.~Xu, H.~Zhao, and J.~Jia, ``Dynamic divide-and-conquer adversarial training
  for robust semantic segmentation,'' in \emph{Proceedings of the IEEE/CVF
  International Conference on Computer Vision}, 2021, pp. 7486--7495.

\bibitem{xiao2018characterizing}
C.~Xiao, R.~Deng, B.~Li, F.~Yu, M.~Liu, and D.~Song, ``Characterizing
  adversarial examples based on spatial consistency information for semantic
  segmentation,'' in \emph{Proceedings of the European Conference on Computer
  Vision (ECCV)}, 2018, pp. 217--234.

\bibitem{JiadongLin2019NesterovAG}
J.~Lin, C.~Song, K.~He, L.~Wang, and J.~E. Hopcroft, ``Nesterov accelerated
  gradient and scale invariance for adversarial attacks,'' \emph{Learning},
  2019.

\bibitem{zhu2022rethinking}
\BIBentryALTinterwordspacing
Y.~Zhu, J.~Sun, and Z.~Li, ``Rethinking adversarial transferability from a data
  distribution perspective,'' in \emph{International Conference on Learning
  Representations}, 2022. [Online]. Available:
  \url{https://openreview.net/forum?id=gVRhIEajG1k}
\BIBentrySTDinterwordspacing

\bibitem{CihangXie2018ImprovingTO}
C.~Xie, Z.~Zhang, Y.~Zhou, S.~Bai, J.~Wang, Z.~Ren, and A.~L. Yuille,
  ``Improving transferability of adversarial examples with input diversity,''
  \emph{computer vision and pattern recognition}, 2018.

\bibitem{YinpengDong2019EvadingDT}
Y.~Dong, T.~Pang, H.~Su, and J.~Zhu, ``Evading defenses to transferable
  adversarial examples by translation-invariant attacks,'' \emph{computer
  vision and pattern recognition}, 2019.

\bibitem{dong2018boosting}
Y.~Dong, F.~Liao, T.~Pang, H.~Su, J.~Zhu, X.~Hu, and J.~Li, ``Boosting
  adversarial attacks with momentum,'' in \emph{Proceedings of the IEEE
  conference on computer vision and pattern recognition}, 2018, pp. 9185--9193.

\bibitem{gu2021adversarial}
\BIBentryALTinterwordspacing
J.~Gu, H.~Zhao, V.~Tresp, and P.~H.~S. Torr, ``Adversarial examples on
  segmentation models can be easy to transfer,'' \emph{CoRR}, vol.
  abs/2111.11368, 2021. [Online]. Available:
  \url{https://arxiv.org/abs/2111.11368}
\BIBentrySTDinterwordspacing

\bibitem{nesterov_accelerated_gradient}
Y.~NESTEROV, ``A method for unconstrained convex minimization problem with the
  rate of convergence $o(1/k^2)$,'' \emph{Doklady AN USSR}, vol. 269, pp.
  543--547, 1983.

\bibitem{momentum_method}
B.~Polyak, ``Some methods of speeding up the convergence of iteration
  methods,'' \emph{Ussr Computational Mathematics and Mathematical Physics},
  vol.~4, pp. 1--17, 12 1964.

\bibitem{agarap2018deep}
\BIBentryALTinterwordspacing
J.~Gu, H.~Zhao, V.~Tresp, and P.~H.~S. Torr, ``Adversarial examples on
  segmentation models can be easy to transfer,'' \emph{CoRR}, vol.
  abs/2111.11368, 2021. [Online]. Available:
  \url{https://arxiv.org/abs/2111.11368}
\BIBentrySTDinterwordspacing

\bibitem{zheng2015improving}
H.~Zheng, Z.~Yang, W.~Liu, J.~Liang, and Y.~Li, ``Improving deep neural
  networks using softplus units,'' in \emph{2015 International joint conference
  on neural networks (IJCNN)}.\hskip 1em plus 0.5em minus 0.4em\relax IEEE,
  2015, pp. 1--4.

\bibitem{he2016deep}
K.~He, X.~Zhang, S.~Ren, and J.~Sun, ``Deep residual learning for image
  recognition,'' in \emph{Proceedings of the IEEE conference on computer vision
  and pattern recognition}, 2016, pp. 770--778.

\bibitem{https://doi.org/10.48550/arxiv.1905.07088}
\BIBentryALTinterwordspacing
Y.~Song, S.~Garg, J.~Shi, and S.~Ermon, ``Sliced score matching: A scalable
  approach to density and score estimation,'' 2019. [Online]. Available:
  \url{https://arxiv.org/abs/1905.07088}
\BIBentrySTDinterwordspacing

\bibitem{nesterov2012efficiency}
Y.~Nesterov, ``Efficiency of coordinate descent methods on huge-scale
  optimization problems,'' \emph{SIAM Journal on Optimization}, vol.~22, no.~2,
  pp. 341--362, 2012.

\bibitem{liu2016delving}
\BIBentryALTinterwordspacing
Y.~Liu, X.~Chen, C.~Liu, and D.~Song, ``Delving into transferable adversarial
  examples and black-box attacks,'' in \emph{5th International Conference on
  Learning Representations, {ICLR} 2017, Toulon, France, April 24-26, 2017,
  Conference Track Proceedings}.\hskip 1em plus 0.5em minus 0.4em\relax
  OpenReview.net, 2017. [Online]. Available:
  \url{https://openreview.net/forum?id=Sys6GJqxl}
\BIBentrySTDinterwordspacing

\bibitem{cai2023ensemblebased}
Z.~Cai, Y.~Tan, and M.~S. Asif, ``Ensemble-based blackbox attacks on dense
  prediction,'' 2023.

\bibitem{pascal-voc-2012}
M.~Everingham, L.~Van~Gool, C.~K.~I. Williams, J.~Winn, and A.~Zisserman, ``The
  {PASCAL} {V}isual {O}bject {C}lasses {C}hallenge 2012 {(VOC2012)}
  {R}esults,''
  http://www.pascal-network.org/challenges/VOC/voc2012/workshop/index.html.

\bibitem{wang2004image}
Z.~Wang, A.~C. Bovik, H.~R. Sheikh, and E.~P. Simoncelli, ``Image quality
  assessment: from error visibility to structural similarity,'' \emph{IEEE
  transactions on image processing}, vol.~13, no.~4, pp. 600--612, 2004.

\bibitem{sandler2018mobilenetv2}
M.~Sandler, A.~Howard, M.~Zhu, A.~Zhmoginov, and L.-C. Chen, ``Mobilenetv2:
  Inverted residuals and linear bottlenecks,'' in \emph{IEEE Conference on
  Computer Vision and Pattern Recognition (CVPR)}, 2018, pp. 4510--4520.

\bibitem{ilyas2019adversarial}
A.~Ilyas, S.~Santurkar, D.~Tsipras, L.~Engstrom, B.~Tran, and A.~Madry,
  ``Adversarial examples are not bugs, they are features,'' 2019.

\bibitem{DBLP:conf/icml/RadfordKHRGASAM21}
A.~Radford, J.~W. Kim, C.~Hallacy, A.~Ramesh, G.~Goh, S.~Agarwal, G.~Sastry,
  A.~Askell, P.~Mishkin, J.~Clark, G.~Krueger, and I.~Sutskever, ``Learning
  transferable visual models from natural language supervision,'' in
  \emph{{ICML}}, ser. Proceedings of Machine Learning Research, vol. 139.\hskip
  1em plus 0.5em minus 0.4em\relax {PMLR}, 2021, pp. 8748--8763.

\bibitem{girdhar2023imagebind}
R.~Girdhar, A.~El-Nouby, Z.~Liu, M.~Singh, K.~V. Alwala, A.~Joulin, and
  I.~Misra, ``Imagebind: One embedding space to bind them all,'' in
  \emph{CVPR}, 2023.

\bibitem{li2022exploring}
\BIBentryALTinterwordspacing
Y.~Li, H.~Mao, R.~B. Girshick, and K.~He, ``Exploring plain vision transformer
  backbones for object detection,'' in \emph{Computer Vision - {ECCV} 2022 -
  17th European Conference, Tel Aviv, Israel, October 23-27, 2022, Proceedings,
  Part {IX}}, ser. Lecture Notes in Computer Science, S.~Avidan, G.~J. Brostow,
  M.~Ciss{\'{e}}, G.~M. Farinella, and T.~Hassner, Eds., vol. 13669.\hskip 1em
  plus 0.5em minus 0.4em\relax Springer, 2022, pp. 280--296. [Online].
  Available: \url{https://doi.org/10.1007/978-3-031-20077-9\_17}
\BIBentrySTDinterwordspacing

\end{thebibliography}
}


\end{document}